\documentclass[
    pdflatex,
    sn-apa, 
    oneside
    ]{sn-jnl}

\usepackage{graphicx}
\usepackage{multirow}
\usepackage{amsmath,amssymb,amsfonts}
\usepackage{amsthm}
\usepackage{mathrsfs}
\usepackage[title]{appendix}
\usepackage{algorithm}
\usepackage{algorithmicx}
\usepackage{algpseudocode}
\usepackage{listings}
\usepackage{float}
\usepackage{etoolbox}
\usepackage{overpic}
\usepackage{url}
\usepackage{xcolor}
\usepackage{textcomp}
\usepackage{manyfoot}
\usepackage{booktabs}
\usepackage{comment}
\usepackage{lscape}
\usepackage{tabularx}
\usepackage{ragged2e}
\usepackage{nameref}
\usepackage{hyperref} 
\usepackage{setspace}
\usepackage{subcaption}

\newgeometry{
    top=1in,
    bottom=1in,
    left=1in,
    right=1in,
    includeheadfoot 
}

\theoremstyle{thmstyleone}
\theoremstyle{thmstyletwo}

\theoremstyle{thmstylethree}

\graphicspath{{figures/}}

\begin{document}

\title[Target-Event-Agent Networks]{The TEA Nets framework combines AI and cognitive network science to model targets, events and actors in text}

\author*[1]{\fnm{Sebastiano} \sur{Franchini}}
\equalcont{These authors contributed equally to this work.}

\author*[1]{\fnm{Alexis} \sur{Carrillo}}
\equalcont{These authors contributed equally to this work.}

\author*[1]{\fnm{Edoardo Sebastiano} \sur{De Duro}}
\equalcont{These authors contributed equally to this work.}

\author*[1]{\fnm{Riccardo} \sur{Improta}}

\author*[1]{\fnm{Ali} \sur{Aghazadeh Ardebili}}

\author*[1,*]{\fnm{Massimo} \sur{Stella}}

\affil*[1]{\orgdiv{CogNosco Lab, Department of Psychology and Cognitive Science, University of Trento}, \city{Rovereto, Trento}, \country{Italy}}

\affil[*]{Corresponding author: massimo.stella-1@unitn.it}

\abstract{
We introduce Target-Event-Agent Networks (TEA Nets) as a computational framework to extract subjects (``Agents"), verbs (``Events"), and objects (``Targets") from texts. Grounded in cognitive network science and artificial intelligence, TEA Nets are implemented as an open-source Python library. We test TEA Nets in three case studies, demonstrating the framework's ability to perform interpretable emotion detection, semantic frame analyses, and linguistic inquiries across conspiracy texts and textual responses generated by LLMs. In the LOCO conspiracy corpus, TEA Nets revealed that highly conspiratorial narratives (4,227 texts) linked personal pronouns (``I", ``you", ``we") with the same actions twice as frequently as low-similarity conspiracy narratives. High-conspiracy narratives connected person-focused elements (``you", ``people") through actions eliciting anger above the random baseline ($z = 2.63, p < .05$), a trend absent in low-similarity conspiracy narratives, which emphasized scientific actors (``researcher", ``scientist"). In the HOPE and CounseLLMe datasets of 212 (human) and 200 (LLM-based) psychotherapy transcripts, respectively, TEA Nets highlighted emotional differences. When expressing feelings, Claude 3 Haiku, GPT-3.5, and humans used sad words with higher frequency than random expectations but Haiku expressed sadness with lower emotional intensity than humans ($U = 1243.5, p = .036$). We discuss these differences in the context of psychotherapy training on LLM-simulated patients. Our results show that Target-Event-Agent Networks can extract relevant emotional, syntactic, and semantic insights from narratives, opening new avenues for text analysis with cognitive network science.}


\maketitle

\section{Introduction}\label{sect:introduction}
Cognitive psychology defines the mental lexicon as a complex dynamical system that acquires, stores, and processes cognitive representations of linguistic concepts \citep{hills2022mind,haim2026cognitive,harnad2017cognize}. Empirical research over the past five decades demonstrates that the mental lexicon operates through associations between concepts \citep{doczi2019overview,aitchison2012words}. Concepts in the mental lexicon are accessed via various associative links, including semantic, syntactic, phonological, orthographic, and emotional associations. The structure of these associations significantly influences cognitive processing, memory, and retrieval efficiency \citep{doczi2019overview,harnad2017cognize,vitevitch2022can}. For instance, the number and diversity of associations connected to a concept directly impact its memorability and retrieval ease \citep{vitevitch2022can}.
Cognitive networks can span multiple dimensions, including semantic, phonological, syntactic, and emotional associations \citep{Semeraro_2025_EmoAtlas,aeschbach2025measuring}. To capture this complexity, multilayer cognitive networks offer a way to simultaneously represent these multiple associative dimensions across concepts \citep{stella2024cognitive}. In a multilayer cognitive network, a concept could be replicated as multiple nodes, one per layer, and linked differently on each and layer. In this way, one concept (e.g., ``love") could be linked with its synonyms on a semantic layer (e.g., with ``passion" or ``affection") and with other words sounding similar to it on a phonological layer (e.g., with ``dove" or ``glove") at the same time. Layers could also be created without replicating nodes, but rather by clustering together nodes of a given type \citep{hills2024behavioral}. This clustering can lead to n-partite networks, i.e., complex networks where nodes are clustered in $n$ groups and links connect only nodes of any two different groups \citep{newman_2018_networks}. Using multiple layers and/or partitions can be crucial for highlighting key structural features of the mental lexicon. For instance, one could partition a cognitive network into several groups, such as verbs, nouns and other parts of speech, and then check their connectivity properties across semantic or syntactic relationships. Before expanding on this research direction, which is key for our Target-Event-Agent Networks framework, let us discuss the dimension of syntactic relationships in cognitive network science.

Traditional cognitive networks often rely on word co-occurrence to extract syntactic relationships between concepts (represented as words) in texts \citep{haim2026cognitive,aitchison2012words}. However, co-occurrence methods can capture spurious associations and may miss meaningful syntactic relationships \citep{aitchison2012words, stella2020text, Teixeira_2021_structure_notes}. For example, in the sentence ``today I feel so much very - well, you know — sick!", the syntactic relationship between ``I" and ``sick" would remain undetected by simple co-occurrence methods due to their textual distance.  
This limitation has motivated cognitive network science to adopt textual forma mentis networks (TFMNs; \citep{stella2022cognitive}).
TFMNs are cognitive networks constructed from textual data that explicitly identify syntactic relationships within sentences, capturing deeper associative structures \citep{stella2020text, stella2022cognitive, Semeraro_2025_EmoAtlas}. This process relies on syntactic parsing by artificial intelligence (AI) models that identify relationships such as subject–verb, verb–object, and noun–adjective structures, representing these connections as edges between concept nodes. Additionally, semantic and emotional properties can enrich TFMNs, creating a rich, multidimensional representation of associative knowledge \citep{stella2022cognitive, Semeraro_2022_Emotional}. This framework has been applied in several domains, including psychopathology in interviews with young adolescents \citep{carrillo2025textual}, creativity in short texts \citep{haim_2024_formamentis}, and simulated LLM dialogues \citep{de2024network}.

Despite their versatility, TFMNs still leave the precise grammatical roles of concepts underspecified, making it hard, for example, to tell who is doing what to whom. Such a gap limits the possibility to reconstruct how people frame actors, actions and consequences in large-scale text analysis. For example, TFMNs do not explicitly distinguish bystanders and perpetrators of sexual assault in online rape narratives \citep{abramski2024voices}. In other words, two different roles expressed with the same concept would be represented as a single node in textual forma mentis networks, leading to ambiguity in reconstructing which syntactic relationships were attributed to a given role. For instance, the sentences ``I was raped by my brother" or ``I was raped, while my brother watched and did nothing" would both feature ``brother" as a node in their TFMNs. However, in one case ``brother" would occupy the role of perpetrator while in the other case ``brother" would fit the role of a bystander.
Given the different psychological perceptions that can be attributed to perpetrators and bystanders in sexual assault narratives \citep{abramski2024voices,labhardt2025intervene}, it is important to have novel network science tools capable of distinguishing among actors, actions, consequences and, overall, framings.

To fill this research gap, the current study introduces a novel cognitive network model: Target-Event-Agent Networks (TEA Nets). This framework explicitly maps cognitive associations into structures composed of subjects (``Agents"), verbs (``Events") and objects (``Targets"). By capturing relational and causal dynamics, the TEA Nets framework extends traditional associative representations, providing deeper insights into cognitive structures underlying narratives. 

\subsection{Natural Language Processing approaches to syntactic relation extraction}
Traditional Natural Language Processing (NLP) approaches do not explicitly and transparently detect syntactic relations from texts. Dictionary-based approaches, including psychological lexicons (i.e., word lists systematically organized around psychological categories such as emotions, cognition, and social processes)
or sentiment lexicons \citep{hutto2014vader}, while highly informative and transparent, fail to capture crucial relational information present in language \citep{feuerriegel2025using}. For example, the Linguistic Inquiry and Word Count \citep{boyd2022liwc, pennebaker2001linguistic, boyd2022development} is a psychologically informed tool that counts meaningful features of texts by simply tracking word appearance, without taking into account who did what and to whom.

Similarly, Bag-of-Words (BoW) models reduce texts to unordered collections of words, hence suffering from semantic information loss \citep{feuerriegel2025using}. Although losing contextual information in a document might not always be an issue (e.g., when counting word frequency), these models are severely limited for tasks requiring deeper understanding, such as detecting cause-and-effect relationships \citep{feder2022causal} or analysing emotional framing related to different characters in different narratives \citep{abramski2024voices}.

More recent distributional models, including BERT-based models \citep{devlin2019bert} and Large Language Models (LLMs), encode words and sentences in high-dimensional vector spaces \citep{binz2023using, mikolov2013efficient}. Through fine-tuning, these models can acquire a degree of understanding regarding agents, actions, and consequences within language \citep{li2023theory}, forming an internal representation of contextual information. However, the mechanisms underlying these representations often remain opaque, making such models function as black boxes \citep{feuerriegel2025using, rudin2019stop, rahwan2019machine}. 

Unlike black-box models, Subject-Verb-Object extraction approaches provide transparent, ``glass-box" access to text analysis, facilitating the reconstruction of networks of actors, actions and consequences as mentioned in narratives. This network approach can merge an NLP pipeline that identifies subjects, verbs, and objects with the complexity science perspective of cognitive networks \citep{Siew_2019_CogNetSci, stella2020text, stella2022cognitive, stella2024cognitive, Teixeira_2021_structure_notes}. 

\subsection{Subject-Verb-Object extraction in NLP}
Subject-Verb-Object (SVO) extraction is a Natural Language Processing (NLP) task involving the identification of core grammatical components of a sentence: the subject (action performer), the verb (action), and the object (entity affected by the action) \citep{Teixeira_2021_structure_notes, Sheikh_2023_causal_networks}. Its goal is to explicitly and transparently represent fundamental relational information within text, clarifying ``who did what to whom", something that dictionary-based and BoW approaches do not explicitly represent, and that transformer models may infer only through less transparent internal representations or task-specific prompting/fine-tuning.
In this context, SVO extraction is a specialized form of relation extraction, i.e., a machine learning task focused on identifying semantic relationships between entities \citep{Detroja_2023_relation_extraction}. SVO targets the fundamental predicate-argument structure, allowing a direct, interpretable mapping of actors, actions, and consequences. 
This explicit structural representation offers insight into agency, causality, and how humans construct narratives, reflecting central aspects of human cognition such as self-understanding and event comprehension \citep{stella2024cognitive}. 
By extracting SVOs from texts, researchers can obtain a nuanced understanding of phenomena like problem framing, attribution of blame, and cognitive biases within narratives \citep{abramski2024voices, reiter_haas_2024_analysis_health_narratives, Sheikh_2023_causal_networks}.
For example, SVO structures have been used in research on causal networks \citep{Sheikh_2023_causal_networks} and analysis of suicide notes \citep{Teixeira_2021_structure_notes}. Recently, \citet{Sheikh_2023_causal_networks} introduced SCANER, a framework transforming raw text into causal networks. Their methodology involves manual text simplification followed by standard pre-processing, leading to automatic Subject-Verb-Object extraction. These structures are subsequently filtered using a 3,000-word dictionary of causal trigger words augmented via WordNet \citep{miller1995wordnet} in order to generate causal networks where subjects are ``cause" nodes, objects are ``effect" nodes, and verbs label relationships \citep{Sheikh_2023_causal_networks}. In another study, \citet{Teixeira_2021_structure_notes}, researchers employed NLP and network science techniques to analyze suicide notes, with a particular focus on self-perceptions and psychological states of the writer.
This approach integrates cognitive data and linguistic benchmarks to extract ideas and emotions, clarifying relationships between concepts and their roles \citep{Teixeira_2021_structure_notes}. However, this analysis relied on an unimodal network approach (i.e., a network composed of nodes of the same type or category) that did not distinguish between subjects, verbs, and objects as separate groups of nodes, thereby limiting the potential for investigating emotional, syntactic, or semantic framing.

\subsection{Overview of the TEA Nets framework}
This manuscript introduces TEA Nets as a novel text analysis framework inspired by cognitive science and based on Artificial Intelligence (AI). These networks map relational information among subjects (``Agents"), verbs (``Events"), and objects (``Targets") within a given text. 
Unlike \citet{Teixeira_2021_structure_notes}, we keep these three components as distinct groups of nodes so that sentences are represented as undirected almost-tripartite graphs, i.e. TEA Nets. In graph theory, tripartite graphs would partition nodes into three groups and feature edges only between nodes belonging to different groups \citep{newman_2018_networks}. Within this novel framework, networks partition nodes as concepts assigned to Agent, Event, or Target groups and also tend to focus on connections between these three categories. However, TEA Nets are almost-tripartite because they also feature relationships between nodes of the same group, i.e. either synonym relationships or concepts having the same role in a given sentence. 

\begin{figure}[h!]
    \centering
    \includegraphics[width=0.65\linewidth]{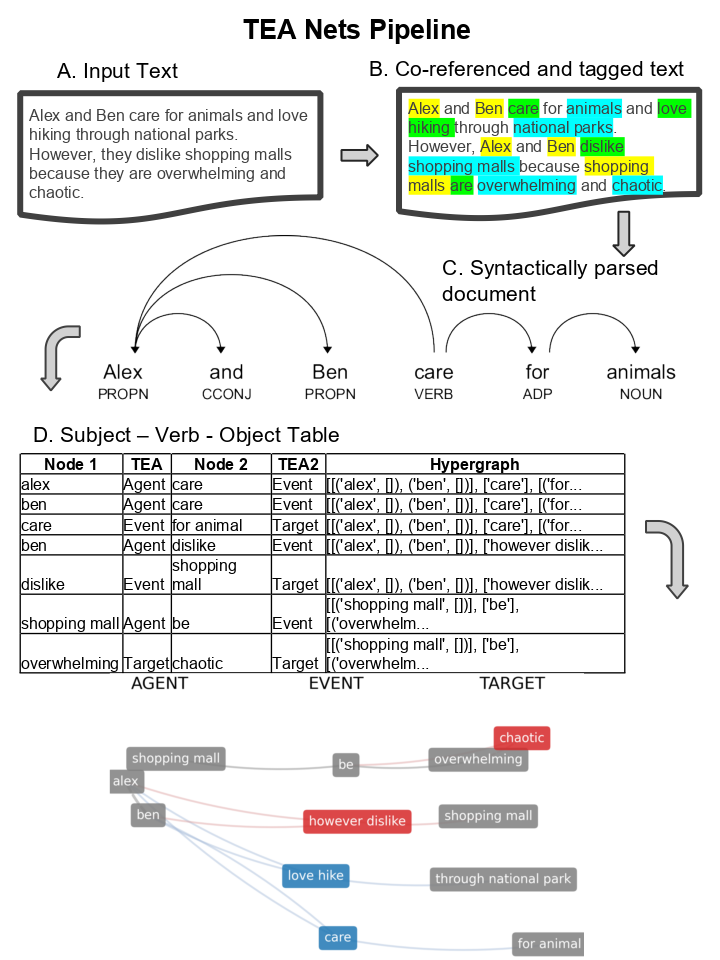}
    \caption{
    Target-Event-Agent Networks (TEA Nets) construction pipeline. This figure illustrates the step-by-step process underlying the TEA Nets framework for extracting relational structure.  Beginning with natural language input (A), the process involves data cleaning and coreference resolution (B), syntactic parsing (C), Subject-Verb-Object (SVO) triplet extraction (D), and produces the final TEA Nets (E).
    }
    \label{fig:infograph_introd}
\end{figure}

Figure \ref{fig:infograph_introd} provides an overview of the TEA Nets construction pipeline. Our methodology processes text by extracting Subject-Verb-Object (SVO; \citep{greenberg1963some, gollob1974subject}), which forms the basis for TEA Nets construction and analysis. SVO tables contain several types of information, including hypergraph descriptions. Hypergraphs are mathematical generalizations of pairwise networks where edges can encompass more than two nodes at a time in the so-called hyperedges \citep{citraro2023hypergraph}. TEA Nets can be considered as hypergraphs when merging together more than two words at a time. For instance, when considering ``terrible tasting wine" as a single entity/node one could: (i) consider ``terrible", ``tasting", ``wine" as separate nodes belonging to the same hyperedge; or (ii) consider ``terrible tasting wine" as aggregated in a single node. Case (i) would correspond to building TEA Nets as semantic hypergraphs. Case (ii) would correspond to visualizing TEA Nets as pairwise networks with nodes containing composite expressions. We believe these two approaches can exploit the power of semantic hypergraph representations of language \citep{citraro2023hypergraph} while simplifying past approaches in the literature \citep{menezes2019semantic}. 

\section{Methodology} \label{sect:methodology} 
Fig. \ref{fig:infograph_introd} provides a visual overview of the Target-Event-Agent Networks (TEA Nets) construction pipeline. The pipeline processes raw text through a sequence of defined stages: text cleaning and coreference resolution (Fig. \ref{fig:infograph_introd}A-B, cf. Section \ref{sect:initial_processing}), syntactic and semantic parsing (Fig. \ref{fig:infograph_introd}C, cf. Section \ref{sec:nlp_parsing}), Subject-Verb-Object (SVO) extraction with passive voice handling (Fig. \ref{fig:infograph_introd}D, cf. Section \ref{sec:svo_extraction}), and network construction (Fig. \ref{fig:infograph_introd}E, cf. Section \ref{sec:network_building}). Section~\ref{sec:measures} introduces the network-level measures adopted throughout our analyses, while Section~\ref{sec:SVO_validation} presents the validation methodology for the SVO extraction.

\textbf{Code Availability.} The Python library implementing Target-Event-Agent Networks is publicly available on GitHub: \url{https://github.com/MassimoStel/TEA_Networks} (Accessed: 28/04/2026), allowing researchers to reproduce and extend our framework across different domains.

\subsection{TEA Nets Construction}

We follow Fig. \ref{fig:infograph_introd} to outline the different stages of construction for TEA Nets starting from a given text.

\subsubsection{Coreference resolution}\label{sect:initial_processing}

Before any structural analysis can take place, the text must be made internally consistent: pronouns and other referring expressions need to be anchored to the entities they stand for. In order to achieve this, the TEA Nets pipeline uses the \texttt{FastCoref} library \citep{Otmazgin2022FcorefFA} as its default resolver. Starting from the raw input (Fig. \ref{fig:infograph_introd}A), the module first groups expressions that likely refer to the same entity into clusters. Then, it identifies any pronouns within these clusters (such as ``they'' or ``it'') and replaces them with the most informative explicit mention available.
We then use the resulting coreferenced text (Fig. \ref{fig:infograph_introd}B) as input for the downstream parsing stages. Because we designed the pipeline to be modular, we can easily integrate alternative coreference resolvers, such as Stanza \citep{qi2020stanza}, without modifying other components.

\subsubsection{Syntactic and Semantic Parsing}\label{sec:nlp_parsing}
The coreferenced text is processed through a standard NLP pipeline built on the Python library \texttt{spaCy} \citep{spacy_website}, utilizing the transformer-based \texttt{en\_core\_web\_trf} model. 

The system performs three sequential operations: first, Part-of-Speech (POS) tagging assigns a grammatical category to each token (e.g., ``Alex'' as a proper noun, ``care'' as a verb). 
Next, dependency parsing maps the syntactic relations between tokens, generating a dependency tree like the one shown in Fig. \ref{fig:infograph_introd}C (where ``Alex'' is the nominal subject of ``care''). 
Finally, lemmatization and Named Entity Recognition (NER) work together to normalize inflected words to their base forms (e.g., ``hiking'' becomes ``hike'') and identify semantic entities such as people and locations.

\subsubsection{Subject-Verb-Object (SVO) extraction}\label{sec:svo_extraction}
Once the document is fully annotated, the pipeline extracts Subject-Verb-Object triples. 
The module first identifies verb candidates (tokens tagged as VERB or AUX whose dependency role is not auxiliary, adjectival, or prepositional) and, for each one, it attempts to recover three components.

\begin{itemize}
    \item[--] Subjects are tokens carrying \texttt{nsubj} or \texttt{nsubjpass} labels; passive constructions are detected by a dedicated module (\S\ref{sec:passive_handling}) that remaps the grammatical subject to its correct semantic role. The extraction function also handles coordinated subjects like ``Alice and Bob'' and cases where a subject must be inherited from an ancestor verb in a conjoined clause, which is a frequent source of errors in simpler extractors.
    \item[--] Verb phrases bundle the main verb lemma with auxiliaries, negations, adverbial modifiers, and any verbs linked via \texttt{xcomp} or \texttt{ccomp} relations. The idea is to capture ``did not seem to care'' as a unit rather than stripping it down to ``care''.
    \item[--] Objects cover direct objects (\texttt{dobj}), prepositional objects (\texttt{pobj}), indirect objects (\texttt{dative}), clausal complements (\texttt{ccomp}, \texttt{xcomp}), and objects inherited from conjoined verbs.
\end{itemize}

Each extracted triplet is stored as pairwise relations (Subject$\to$Verb, Verb$\to$Object) in a structured DataFrame (Fig. \ref{fig:infograph_introd}D), with additional hypergraph-level information linking all three elements of the same triplet. Each row carries two binary flags related to the voice of the verb. The first flag, \texttt{is\_passive}, simply records whether the verbal construction is passive (1) or active (0). The second flag, \texttt{passive\_approx}, equals~1 when the verb is passive but no explicit agent (\textit{by}-phrase) was found in the text; in these cases the grammatical subject (i.e.\ the patient) is placed in the \textit{Agent} slot as the best available approximation rather than being discarded. Downstream analyses can filter on these flags to distinguish genuine agentive relations from approximated ones or to study the overall usage of passive voice (\S\ref{sec:passive_handling}).

\subsubsection{Passive Voice Handling}\label{sec:passive_handling}
Passive voice is one of the most challenging aspects of SVO extraction, and misidentifying it has significant consequences for narrative analysis: treating the grammatical subject as the semantic agent can turn victims into perpetrators and vice versa~\citep{abramski2024voices}. The pipeline detects passive constructions through four independent signals, evaluated by a dedicated function:

\begin{itemize}
    \item[--] \textbf{Canonical passive.} The verb carries dependents labeled \texttt{auxpass} (or \texttt{aux:pass}) and/or \texttt{nsubjpass} (\texttt{nsubj:pass}). This covers standard \textit{be}-passives (``I was raped'') as well as \textit{get}-passives (``He got arrested''), which \texttt{en\_core\_web\_trf} analyzes identically.
    \item[--] \textbf{Feel-passive.} The verb is a past participle (POS tag \texttt{VBN}) and has a child with \texttt{dep=aux} whose lemma is \emph{not} a modal or copula (e.g.\ ``I felt abused'', ``She seemed targeted''). In these constructions \texttt{spaCy} does not use \texttt{auxpass}, so the canonical check alone would miss them. We maintain a whitelist of non-passive auxiliary lemmas (\textit{be, have, do, will, shall, would, should, could, might, may, must, can}) and treat any other auxiliary before a \texttt{VBN} as a passive signal.
    \item[--] \textbf{Agent-phrase.} The verb is \texttt{VBN} and has a child with \texttt{dep=agent} (a \textit{by}-phrase), regardless of the auxiliary type. This catches cases like ``He appeared beaten by the crowd'', where the parser treats ``appeared'' as the main verb and ``beaten'' as an \texttt{xcomp} or \texttt{oprd} complement.
    \item[--] \textbf{Coordinated conjunct.} The verb is \texttt{VBN} with \texttt{dep=conj} and its head verb is itself passive (as determined recursively by the same function). This handles sentences such as ``I was held down and raped'', where ``raped'' has no auxiliary of its own but shares the passive frame of ``held''. If the head verb has an explicit agent, it is propagated to the conjunct.
\end{itemize}

Once a verb is identified as passive, the algorithm follows one of two paths. If an explicit \textit{by}-phrase is present (the dependency path \texttt{agent}~$\to$~\texttt{pobj}), the object of that preposition is assigned to the \textit{Agent} role as the semantic actor, while the grammatical subject is moved to \textit{Target}. For example, ``I was raped by my uncle'' maps to \textit{Agent}~=~uncle, \textit{Event}~=~rape, \textit{Target}~=~I. The same logic applies to coordinated passives such as ``The report was written and reviewed by the committee'', where the agent may attach to a conjoined verb rather than the root.

When no \textit{by}-phrase is present, the semantic agent is unobserved. The current implementation retains the grammatical subject as an approximated placeholder and marks the row with \texttt{passive\_approx\,=\,1}; however, downstream analysis of agency should either exclude these rows or treat the \textit{Agent} value as an unresolved patient placeholder rather than as a true semantic agent (with \texttt{is\_passive} set to 1). This flag allows downstream analyses to distinguish genuine agentive relations from approximated ones: for instance, a perpetrator-frequency analysis can exclude rows with \texttt{passive\_approx\,=\,1} to avoid counting victims as agents, while a victim-exposure analysis can select precisely those rows to quantify how often assault events are narrated without naming a perpetrator. Furthermore, having an explicit \texttt{is\_passive} flag for all rows allows researchers to easily aggregate and compare the prevalence of active versus passive narrative framing across the corpus.

Finally, for any verb flagged as passive, the subject-extraction function returns immediately, preventing the inheritance mechanism from climbing the dependency tree and erroneously assigning an active ancestor's subject to the passive verb. Without this guard, a sentence like ``I went out and got raped'' would assign ``I'' as the active agent of ``raped'', because the verb ``went'' (an active ancestor) carries ``I'' as its \texttt{nsubj}.

\subsubsection{Network dataframe and node/links attributes}\label{sec:network_building}

Subjects, verbs, and objects become nodes labelled with their respective roles (\textit{Agent}, \textit{Event}, \textit{Target}), forming a tripartite graph. Undirected edges connect subjects to verbs and verbs to objects. The visual representation of the network encodes several layers of information: edge thickness maps to the weight of the syntactic relations, while edge colors indicate the sentiment interaction between the connected nodes. Each node receives a sentiment polarity score (positive, negative, or neutral) from the VADER lexicon \citep{hutto2014vader}, which determines its color. We chose VADER over learned sentiment models because its rule-based design makes the scoring easy to interpret and reproduce. Intra-layer edges, shown as dashed lines in the figures, link nodes of the same syntactic role and can reflect coreference associations or semantic relations such as synonymy via WordNet \citep{miller1995wordnet}. Finally, relations marked with \texttt{passive\_approx} can be visually distinguished with dotted lines, separating approximated victim-in-Agent relations from genuine agentive edges.

\subsection{Measures for text analysis with TEA Nets}\label{sec:measures}

An important advantage of turning textual unstructured data into a network is relative to being able to apply several network measures for the investigation of texts. We define here a set of quantities for TEA Nets-based analysis. These quantities are not exhaustive, but they provide a concrete starting point.\\

\textbf{Degree} ($K$) counts the distinct edges connecting a node to nodes of a different type. For an \textit{Agent} node, $K$ is the number of unique \textit{Event}-nodes it links to.\\

\textbf{Relative degree} normalizes degree within a role category:
\begin{equation}
    K^*_i = \frac{K_i}{\sum_{j \neq i} K_j}.
\end{equation}\\

\textbf{Lemma frequency} ($F$) records the total occurrences of a lemma across all relevant edges, regardless of syntactic function.\\

\textbf{Repetitiveness Index \textit{(RI)}} measures how repetitively a lemma is used:
\begin{equation}
    \mathrm{RI}_i = \frac{F_i}{K_i}.
\end{equation}
High values of \textit{RI} mean that the same lemma appears repeatedly in a small number of connections, whereas low values reflect a more varied structural use.\\

\textbf{Edge weight} measures how often a specific syntactic relation occurs (e.g., \textit{Ben}--\textit{dislike}). The normalized version,
\begin{equation}
    \mathrm{NW}_{e,S} = \frac{F_{e,S}}{\sum_{e' \in E_{\mathrm{type},S}} F_{e',S}},
\end{equation}
allows comparison across sub-corpora of different sizes.\\

\textbf{Prominence} measures how much more (or less) central an edge is in one sub-corpus with respect to another:
\begin{equation}\label{eq:edge prominence}
    P_e = \mathrm{NW}_{e,1} - \mathrm{NW}_{e,2}.
\end{equation}

\subsection{Validation of the SVO Extraction Pipeline}\label{sec:SVO_validation}

For validation, we built a benchmark of 122 English sentences designed to stress-test the extraction module on the syntactic phenomena most likely to cause problems. The first 100 sentences form a general benchmark covering simple active clauses, passive constructions with and without an explicit agent, progressive and perfect tenses, modals, negation, yes/no and wh- interrogatives, relative clauses, coordinated subjects and objects, double-object constructions, infinitival complements (\texttt{xcomp}), adverbial subordinates, intransitive verbs, and existential constructions. The remaining 22 sentences target passive voice specifically, including canonical \textit{be}-passives, \textit{get}-passives (e.g.\ ``He got arrested by the police''), feel-passives with non-standard auxiliaries (e.g.\ ``I felt abused by him'', ``She seemed targeted''), coordinated passive conjuncts (e.g.\ ``I was held down and raped''), and negative controls consisting of active sentences with superficially similar structure (e.g.\ ``She got lucky'', ``He felt happy''). 
After excluding one ambiguous sentence, the final benchmark contains 122 sentences: 71 active and 51 passive, of which 23 include an explicit \textit{by}-agent and 27 are agentless.

Each sentence was manually annotated with its gold-standard \textit{Agent}, \textit{Event} (lemmatized), and \textit{Target} semantic roles following the TEA convention described in \S\ref{sec:svo_extraction}--\ref{sec:passive_handling}. For active constructions, the grammatical subject maps to \textit{Agent} and the object to \textit{Target} (e.g.\ ``The cat chased the mouse'' $\to$ \textit{Agent}~=~cat, \textit{Event}~=~chase, \textit{Target}~=~mouse). For passive constructions with an explicit agent, the \textit{by}-phrase argument maps to \textit{Agent} and the grammatical subject to \textit{Target} (e.g.\ ``The mouse was chased by the cat'' $\to$ \textit{Agent}~=~cat, \textit{Event}~=~chase, \textit{Target}~=~mouse). For agentless passives, the grammatical subject is kept in \textit{Agent} as a best-available approximation with the \texttt{passive\_approx} flag set to~1 (e.g.\ ``The window was broken'' $\to$ \textit{Agent}~=~window, \textit{Event}~=~break, \textit{Target}~=~$\varnothing$). The benchmark and the annotation notebook are released with the library.

We compare two extractors against the gold annotations.

\textbf{Baseline.} A minimal dependency-based extractor that processes one sentence at a time. For each sentence, it locates the ROOT verb in the dependency tree produced by \texttt{spaCy} and collects the grammatical subject (\texttt{nsubj} or \texttt{nsubjpass} dependent) and the first available object (\texttt{dobj}, \texttt{pobj}, \texttt{attr}, \texttt{acomp}, \texttt{ccomp}, or the object of a prepositional or \texttt{xcomp} complement). When the subject carries the \texttt{nsubjpass} label or the verb has an \texttt{agent} dependent (\textit{by}-phrase), the extractor applies the same syntactic-to-semantic mapping used in the gold annotations: in passive constructions with an explicit agent the object of the \textit{by}-phrase becomes the predicted \textit{Agent} and the grammatical subject becomes the predicted \textit{Target}; in agentless passives the grammatical subject is placed in the \textit{Agent} slot. In active constructions the grammatical subject maps to \textit{Agent} and the object to \textit{Target}. This baseline isolates the contribution of \texttt{spaCy}'s dependency parser without any of the additional heuristics (verb-phrase bundling, subject inheritance, coordinated-object handling) introduced by the full TEA pipeline.

\textbf{TEA Nets.} The full extraction pipeline described in sections \ref{sec:svo_extraction}--\ref{sec:passive_handling}. For each sentence, \texttt{extract\_svos()} returns a relational DataFrame from which the first \textit{Agent$\to$Event} and \textit{Event$\to$Target} edges are taken as the predicted triple.

\textbf{Performance Measurements.} Accuracy is computed separately for each semantic role (\textit{Agent}, \textit{Event}, \textit{Target}) by comparing the predicted lemma against the gold lemma after lowercasing and treating empty predictions and empty gold values as \texttt{\_\_none\_\_}. True positives are counted only when both prediction and gold are non-empty and match.

\textbf{Passive Voice Validation.} To evaluate the passive-handling algorithm independently from role assignment, we use the crowd-sourced dataset released with PassivePy \citep{sepehri2023passivepy}. This dataset comprises 1{,}150 English sentences drawn from diverse online sources, each independently annotated for passive voice by human raters via Amazon Mechanical Turk. Of these, 216 sentences (18.8\%) are labeled as passive and 934 (81.2\%) as active. For each sentence, the TEA pipeline's \texttt{\_passive\_info()} function is applied to every verb candidate, and the sentence is predicted as passive if at least one verb is flagged. We report per-class Correct, Total, and Accuracy for both the Passive and Active classes, together with the macro average.

\subsubsection{Validation Results}\label{sec:validation_results}

Table~\ref{tab:validation_results} reports accuracy for both the baseline extractor and TEA Nets across the three semantic roles on the 122-sentence benchmark. Table~\ref{tab:passive_validation_results} reports passive voice detection performance on the 1{,}150-sentence PassivePy crowd-sourced dataset.

\begin{table}[hbpt!]
\centering
\begin{minipage}[t]{0.54\textwidth}
    \centering
    \caption{Accuracy for the baseline SVO extractor and TEA Nets on the 122-sentence benchmark, evaluated separately for each semantic role.}
    \label{tab:validation_results}
    \begin{tabular}{@{}llccc@{}}
    \toprule
    Extractor & Role & Correct & Total & Accuracy\\
    \midrule
    Baseline & Agent & 109 & 122 & 0.893\\
    Baseline & Event & 85 & 122 & 0.697\\
    Baseline & Target & 96 & 122 & 0.787\\
    TEA Nets & Agent& 111 & 122 & \textbf{0.910}\\
    TEA Nets & Event & 99 & 122 & \textbf{0.811}\\
    TEA Nets & Target & 105 & 122 & \textbf{0.861}\\
    \bottomrule
    \end{tabular}
\end{minipage}\hfill
\begin{minipage}[t]{0.42\textwidth}
    \centering
    \caption{Passive voice detection evaluated against the PassivePy crowd-sourced dataset (1{,}150 sentences with human annotations).}
    \label{tab:passive_validation_results}
    \begin{tabular}{@{}lccc@{}}
    \toprule
    Class & Correct & Total & Accuracy \\
    \midrule
    Passive   & 206 & 216 & 0.954 \\
    Active    & 911 & 934 & 0.975 \\
    \midrule
    Macro Avg & 1117 & 1150 & 0.971 \\
    \bottomrule
    \end{tabular}
\end{minipage}
\end{table}


\section{Case studies}\label{sect:use_cases} 
The TEA Nets framework offers a powerful and comprehensive approach for narrative analysis by systematically identifying entities, interactions, and associated emotional valences. We demonstrate its practical utility and versatility within three distinct case studies. 

In Section \ref{sect:conspiracy_narrative} we present an analysis of high- and low-similarity conspiracy documents drawn from the LOCO corpus \citep{Miani2022LOCO}. In Section \ref{sect:assault} we use the TEA Nets Python library to process and analyze a corpus of Reddit posts and comments discussing sexual assault, focusing on how agents are linguistically represented in active and passive constructions. In Section  \ref{sect:CounseLLMe} we apply TEA Nets to a comparative investigation of counseling dialogues involving both human patients and Large Language Model-enhanced virtual patients \citep{deduro2025counsellme}.


\subsection{Case 1: Analysis of psychological social media posts about conspiracy theory} \label{sect:conspiracy_narrative}
The Target-Event-Agent Networks framework offers a useful approach for information extraction from large corpora, including those involving conspiracy narratives. Traditional NLP methods, specifically those that process surface-level linguistic data (e.g., dictionary-based and Bag-of-Words models), exhibit limitations with syntactic dependencies and the hierarchical organization of information \citep{Kenett_2022_Networks, Semeraro_2025_EmoAtlas}. Furthermore, these methods fail to extract information regarding claim attribution or implicit causal link establishment \citep{Sheikh_2023_causal_networks}.
By explicitly modeling causality and agency, the TEA Nets framework is suited to conspiracy-narrative analysis by extracting structured information that enables formal grammar investigation and delineation of main roles within conspiracy narratives \citep{reiter_haas_2024_analysis_health_narratives}. 
We implemented a case study designed to illustrate the capabilities of Target-Event-Agent Networks. In order to do so, we employed a dedicated dataset of conspiracy documents (Language of Conspiracy; \citep{Miani2022LOCO}). 
It is important to note that this case study is intended to be illustrative only and does not constitute an exhaustive analysis of conspiracy texts.
For broader investigations, see \citep{gagliardi_2023_conspiracy_bias_beliefs, Gambini_2024_anatomy_conspiracy_theorists}.

\subsubsection{Dataset}
To illustrate the utility of the TEA Nets, we adopted the LOCO corpus \citep{Miani2022LOCO} for the study of conspiracy theories. 
This corpus comprises $96,743$ standalone documents extracted from various websites. Our analysis focuses on the $23,937$ documents within the conspiracy subcorpus, gathered from websites specifically identified as promoting conspiracy content.
For our comparative analysis, we divided the conspiracy subcorpus into two distinct subsets based on the \texttt{cosine\_similarity} metadata field. This information makes it possible to measure how closely the language of each document matches the average linguistic profile of all conspiracy documents in the subcorpus, using cosine similarity. Cosine similarity values exceeding one standard deviation above the mean were flagged as ``conspiracy representative" in the dataset metadata. Following this definition, our ``high-conspiracy documents" subset comprised the $4,227$ texts flagged as ``conspiracy representative" ($cosine\_similarity > mean + SD$). Conversely, for the ``low-similarity conspiracy documents" another $4,227$ texts were selected from those documents exhibiting the lowest cosine similarity values within the dataset. This approach enables a direct comparison between documents that are most and least characteristic of the conspiracy discourse. Since this selection process accentuates differences, it minimizes overlap between the high-conspiracy and low-similarity conspiracy groups, providing a basis for identifying linguistic and thematic distinctions within the conspiracy subcorpus.
To ensure consistency and mitigate confounding variables, we prioritized topic matching over the use of diverse datasets. While aligning thematic content across different sources presents significant methodological challenges, our approach successfully maintained this coherence. Future research should focus on the curated construction of balanced datasets specifically designed for topic matching between conspiratorial and non-conspiratorial narratives.

\subsubsection{Methods}
For each of the $4,227$ texts, the \texttt{teanets.extract\_svos\_from\_text()} function was used to extract SVOs and create an SVO \texttt{DataFrame}. This procedure was systematically applied to all documents in our corpus, with each document's SVOs being saved in its respective \texttt{DataFrame}. Subsequently, we used the \texttt{teanets.analytics.merge\_svo\_dataframes()} to aggregate the individual SVOs into a single unified \texttt{DataFrame} for each set of documents (i.e., high-conspiracy and low-similarity conspiracy categories).

We focused on examining the aggregated SVO data using the functions provided with the TEA Networks Python library. 
We computed network-based measures (see Section \ref{sec:measures}), such as node degree, to quantify and characterize the syntactic relationships embedded within the texts.
To analyze differences between high-conspiracy and low-similarity conspiracy documents, normalized edge weight served as the basis for two comparative analyses: Kendall's $\tau$ correlation and prominence analysis.
To assess the commonality of the ranking of verb usage for a given subject across these two document types, Kendall's $\tau$ correlation was employed.
For this measure, ``Agent-Event" edges connected to a specific ``Agent" node of interest were selected from both high-conspiracy and low-similarity conspiracy documents. The normalized edge weight was retrieved for each of these selected edges in both document sets. For example, considering the ``I" pronoun as an ``Agent" node, the edge ``I-think" had a normalized edge weight of $0.0078$ in high-conspiracy documents and $0.0019$ in low-similarity conspiracy documents. Other verbs connected to ``I" in both document sets included ``attend" and ``assume". In total, 626 such shared ``Agent-Event" edges were identified for the ``I" pronoun. Kendall's $\tau$ was then calculated on the normalized edge weights of these 626 shared edges.

Subsequently, prominence analysis calculated an edge's prominence as the difference between its normalized edge weight in high-conspiracy documents and low-similarity conspiracy documents (as detailed in Section \ref{sec:measures}, Equation \ref{eq:edge prominence}).
For example, if the ``I-think" (``Agent-Event") edge had a normalized edge weight of $.0078$ in high-conspiracy documents and $.0019$ in low-similarity conspiracy documents, a difference of $.0059$ indicates this edge was more prominent in the high-conspiracy narrative.
For visualization, the \texttt{teanets.teaplot.plot\_svo\_graph()} was employed to illustrate the relationships between subjects, verbs, and objects. To enrich the representation, we paired the network with ``emotional flowers" as developed by \citet{Semeraro_2025_EmoAtlas} for each class of documents (conspiracy and low-similarity conspiracy categories). These visualizations summarize the emotions expressed in texts by depicting the z-scores associated with each emotion (as elicited from a text) tested against a random baseline. Details on EmoAtlas implementation and rationale can be found in \citet{Semeraro_2025_EmoAtlas}.

\subsubsection{Results}
\noindent \textbf{\textit{``You", ``I" and ``we" highlight repetitive conspiratorial narratives.}}\\
\noindent To investigate the content of conspiracy narratives, we employed Degree ($K$), Lemma Frequency ($F$), and Repetitiveness Index ($RI$) analysis (see Section \ref{sec:measures}). Table \ref{tab:nodes_degree_loco} presents the top 20 ``Agent" and ``Event" nodes for both high- and low-similarity conspiracy documents within the LOCO dataset. In general, across both node types (``Agent" and ``Event"), absolute frequency and degree values are higher in high-conspiracy documents compared to low-similarity conspiracy documents.

\begin{table}[hbpt!]
\centering
\caption{The degree ($K$), lemma frequency ($F$), and Repetitiveness Index ($RI = F/K$) for the top 20 ``Agent" (subject) and ``Event" (verb) nodes in the LOCO dataset are presented, identified from Subject-Verb-Object (SVOs) derived from both high- and low-similarity conspiracy documents. All degree and frequency values are expressed in thousands.}
\label{tab:nodes_degree_loco}
\setlength{\tabcolsep}{4pt}
\begin{tabular}{@{}lrrrlrrr|lrrrlrrr|@{}}
\toprule
\multicolumn{8}{c|}{High-Conspiracy} & \multicolumn{8}{c|}{Low-similarity conspiracy} \\ \midrule
\multicolumn{1}{c}{Agent} & \multicolumn{1}{c}{K} & \multicolumn{1}{c}{F} & \multicolumn{1}{c|}{RI} & \multicolumn{1}{c}{Event} & \multicolumn{1}{c}{K} & \multicolumn{1}{c}{F} & \multicolumn{1}{c|}{RI} & \multicolumn{1}{c}{Agent} & \multicolumn{1}{c}{K} & \multicolumn{1}{c}{F} & \multicolumn{1}{c|}{RI} & \multicolumn{1}{c}{Event} & \multicolumn{1}{c}{K} & \multicolumn{1}{c}{F} & \multicolumn{1}{c|}{RI} \\ \midrule
you & 18.1 & 56.6 & \multicolumn{1}{r|}{3.1} & be & 77.7 & 418.9 & 5.4 & you & 2.6 & 3.4 & \multicolumn{1}{r|}{1.3} & be & 12.8 & 58.6 & 4.6 \\
we & 17.6 & 69.5 & \multicolumn{1}{r|}{3.9} & have & 13.9 & 114.6 & 8.3 & we & 1.8 & 4.1 & \multicolumn{1}{r|}{2.2} & say & 4.2 & 5.0 & 1.2 \\
I & 16.8 & 64.9 & \multicolumn{1}{r|}{3.9} & say & 12.2 & 25.1 & 2.1 & it & 1.6 & 10.9 & \multicolumn{1}{r|}{6.8} & have & 2.6 & 16.5 & 6.3 \\
it & 9.0 & 89.0 & \multicolumn{1}{r|}{9.9} & make & 6.6 & 19.8 & 3.0 & I & 1.1 & 2.4 & \multicolumn{1}{r|}{2.1} & find & 1.6 & 2.3 & 1.5 \\
they & 4.9 & 79.9 & \multicolumn{1}{r|}{16.3} & see & 6.5 & 17.6 & 2.7 & that & 0.9 & 20.5 & \multicolumn{1}{r|}{22.4} & show & 1.4 & 2.2 & 1.6 \\
people & 4.9 & 31.6 & \multicolumn{1}{r|}{6.5} & know & 5.8 & 21.1 & 3.7 & people & 0.6 & 2.8 & \multicolumn{1}{r|}{4.6} & use & 1.4 & 4.0 & 2.9 \\
that & 4.6 & 150.7 & \multicolumn{1}{r|}{32.7} & take & 5.6 & 16.1 & 2.9 & who & 0.6 & 4.1 & \multicolumn{1}{r|}{7.3} & include & 1.3 & 2.2 & 1.6 \\
who & 3.8 & 37.2 & \multicolumn{1}{r|}{9.7} & call & 5.6 & 12.3 & 2.2 & which & 0.5 & 4.7 & \multicolumn{1}{r|}{8.8} & make & 1.3 & 3.0 & 2.3 \\
those & 3.0 & 13.2 & \multicolumn{1}{r|}{4.4} & be not & 5.6 & - & - & those & 0.4 & 1.4 & \multicolumn{1}{r|}{3.1} & take & 1.3 & 2.3 & 1.9 \\
this & 2.3 & 75.1 & \multicolumn{1}{r|}{32.7} & use & 5.2 & 19.1 & 3.7 & researcher & 0.4 & 0.9 & \multicolumn{1}{r|}{2.6} & report & 1.1 & 2.9 & 2.6 \\
which & 2.1 & 27.6 & \multicolumn{1}{r|}{12.9} & tell & 4.7 & 10.2 & 2.2 & scientist & 0.4 & 1.0 & \multicolumn{1}{r|}{2.6} & tell & 1.0 & 1.0 & 1.0 \\
one & 2.1 & 30.2 & \multicolumn{1}{r|}{14.3} & become & 4.6 & 9.3 & 2.0 & they & 0.4 & 5.2 & \multicolumn{1}{r|}{14.3} & call & 1.0 & 1.5 & 1.6 \\
government & 2.0 & 20.6 & \multicolumn{1}{r|}{10.1} & give & 4.6 & 11.0 & 2.4 & this & 0.4 & 9.7 & \multicolumn{1}{r|}{26.7} & see & 0.9 & 1.9 & 2.0 \\
trump & 1.9 & 9.3 & \multicolumn{1}{r|}{5.0} & get & 4.3 & 15.0 & 3.5 & company & 0.3 & 1.4 & \multicolumn{1}{r|}{4.4} & give & 0.8 & 1.2 & 1.5 \\
man & 1.8 & 9.4 & \multicolumn{1}{r|}{5.1} & include & 4.3 & 8.7 & 2.1 & one & 0.3 & 3.9 & \multicolumn{1}{r|}{13.0} & cause & 0.8 & 2.1 & 2.6 \\
all & 1.8 & 38.8 & \multicolumn{1}{r|}{21.1} & do & 4.1 & 53.7 & 13.1 & some & 0.3 & 2.1 & \multicolumn{1}{r|}{7.4} & know & 0.7 & 1.9 & 2.5 \\
god & 1.6 & 10.3 & \multicolumn{1}{r|}{6.2} & go & 3.9 & 20.8 & 5.4 & group & 0.3 & 1.4 & \multicolumn{1}{r|}{5.1} & be not & 0.7 & - & - \\
some & 1.6 & 14.3 & \multicolumn{1}{r|}{8.9} & think & 3.9 & 10.1 & 2.6 & study & 0.3 & 2.6 & \multicolumn{1}{r|}{9.6} & provide & 0.7 & 0.9 & 1.3 \\
what & 1.6 & 30.8 & \multicolumn{1}{r|}{19.2} & find & 3.8 & 9.5 & 2.5 & what & 0.3 & 2.2 & \multicolumn{1}{r|}{8.4} & add & 0.7 & 0.9 & 1.3 \\
anyone & 1.3 & 3.0 & \multicolumn{1}{r|}{2.2} & come & 3.8 & 14.3 & 3.8 & woman & 0.3 & 0.9 & \multicolumn{1}{r|}{3.6} & write & 0.7 & 0.7 & 1.0 \\ 

\bottomrule
\end{tabular}
\end{table}

The analysis of the ``Agent" nodes reveals distinct patterns in subject representation across high- and low-similarity conspiracy document types within the LOCO dataset.  
As reported in Table \ref{tab:nodes_degree_loco}, in high-conspiracy documents, ``you" ($K=18.1$, $F=56.6$), ``we" ($K=17.6$, $F=69.5$), and ``I" ($K=16.8$, $F=64.9$) all exhibit high and comparable degrees, i.e. a large number of connections with several action (``Event") nodes. These larger degrees indicate the prominence of these pronouns in collective narratives, i.e. ``Agent" nodes with larger degrees are linked with more actions and thus are prominent in the narratives encoded within the corpus. Conversely, in low-similarity conspiracy documents, these pronouns show lower degrees and frequencies (``you": $K=2.6$, $F=3.4$; ``I": $K=1.1$, $F=2.4$; ``we": $K=1.8$, $F=4.1$). The Repetitiveness Index ($RI$) values for first and second-person pronouns in high-conspiracy documents are: ``I" ($RI=3.9$), ``you" ($RI=3.1$) and ``we" ($RI=3.9$). This means that, for instance, ``I" is mentioned roughly four times in connection with the same action (verb) node. In low-similarity conspiracy documents, $RI$ values are lower (``I": $RI=2.1$, ``you": $RI=1.3$, ``we": $RI=2.2$). These patterns indicate that high-conspiracy documents use "you", "I", and "we" in relation to the same actions roughly twice as often as low-similarity conspiracy documents, reflecting a more constrained and repetitive style.\\

\noindent \textbf{\textit{The elevated RI for ``I" in high-conspiracy texts may reflect a pronounced self-referential linguistic pattern. }}\\
\noindent The Repetitiveness Index (RI) for the ``Agent" node ``I" is $3.9$ in high-conspiracy documents, almost double the $2.1$ observed in low-similarity conspiracy documents. This difference is noteworthy considering TEA Nets' lemmatisation process, which consolidates variations of ``I" (e.g., me, myself) to their root form for analysis. The elevated RI for ``I'' in high-conspiracy texts may therefore reflect a pronounced self-referential linguistic pattern, consistent with prior findings on the use of reflexive and possessive pronouns in conspiracy discourse \citep{recordare_2024_conspiracy_theorists}.\\

\noindent \textbf{\textit{The ``Agent" node ``god" is prevalent in high-conspiracy documents.}}\\
\noindent Among the most central ``Agent" nodes in high-conspiracy documents, ``god" ranks 16th, exhibiting a degree of $1,642$ and a frequency of $10,260$. Its Repetitiveness Index was $6.24$. Conversely, in low-similarity conspiracy documents, ``god" ranks 1339th, with a degree of $12$, an edge frequency of $376$, and a Repetitiveness Index of $31.33$. Lower Repetitiveness Index, coupled with higher degree, indicates ``god" connects to a broader range of actions within high-conspiracy documents rather than forming concentrated associations. The prominence of ``god" in high-conspiracy narratives can be attributed to several interconnected factors. First, conspiracy narratives often refer to religious ideas and motifs, providing a foundation for the term's prominence in this discourse \citep{Fong_2021_language_conspiracy}. Second, the increased use of profanity and emotionally charged language in conspiracy narratives \citep{Gambini_2024_anatomy_conspiracy_theorists, Giachanou2023consp_propagators_linguistic} results in the frequent appearance of the word "god" within expletive phrases, thereby incorporating this concept into a wide array of reactive or emotional expressions. 
Taken together, this peculiar usage of the word "god" in high-conspiracy documents may account for the node's numerous yet widely distributed connections, resulting in a lower Repetitiveness Index despite its high overall degree.\\

\noindent \textbf{\textit{Despite limited overlap, TEA Nets reveal distinct verb associations on ``Agent-Event" edges.}}\\
\noindent Out of $909,184$ ``Agent-Event" edges extracted from both high- and low-similarity conspiracy documents, only 1.98\% (18,256) are present in both subcorpora.
This low overlap indicates distinct linguistic patterns between high- and low-similarity conspiracy documents \citep{Giachanou2023consp_propagators_linguistic}. To explore this overlap, correlation and prominence analyses were conducted on the shared subset.
Kendall's $\tau$ correlation values for ``Agent" nodes ``I" ($\tau = 0.458$, $p < .001$), ``you" ($\tau = 0.440$, $p < .001$), and ``we" ($\tau = 0.463$, $p < .001$) indicate that verbs highly frequent with these pronouns in one document type tend to maintain a similarly high rank in the other.
However, these correlations are notably less than one, suggesting relevant qualitative differences in verb usage despite shared high-ranking patterns.
To further explore these observed differences in ``Agent-Event" edges, prominence analysis was performed. Table \ref{tab:who-did_edges_iyw} reports the top 20 prominent ``Agent-Event" edges. The ``Agent" nodes ``I", ``you", and ``we" were selected for this prominence analysis due to their usage across the subcorpora and their differential Repetitiveness Index. By analyzing the prominence, we found a distinct link between pronouns and actions. High-conspiracy documents tend to link these pronouns with verbs expressing conviction (``think", ``know", ``believe"), and conveying action (``do", ``get", ``call", ``bring", ``make", ``say"). Low-similarity conspiracy documents, conversely, use verbs emphasizing information exchange (``search", ``read", ``uncover", ``recommend") and transactional engagement (``order", ``follow", ``download").\\

\begin{table}[ht!]
\centering
\caption{20 most prominent edges connecting ``I", ``you", and ``we" pronouns as ``Agent" nodes to ``Event" nodes for high-conspiracy documents (HC) and low-similarity conspiracy documents (LC).}
\label{tab:who-did_edges_iyw}
\footnotesize
\begin{tabular}{@{}lll|ll|ll|@{}}
\toprule
Pronoun & \multicolumn{2}{c|}{I} & \multicolumn{2}{c|}{you} & \multicolumn{2}{c|}{we} \\ \midrule
Document Set & HC & LC & HC & LC & HC & LC \\ \midrule
Rank &  &  &  &  &  &  \\ \cmidrule(r){1-1}
1 & be & stay & be & order today & be & must make \\
2 & think & 've publish & know & can follow & have & 've see \\
3 & know & 've work & let & check & know & halt \\
4 & have & 've tell & see & note & see & discuss \\
5 & say & will receive & do & visit & do & collectively face \\
6 & believe & add & think & download & present & 'll look \\
7 & see & do not recommend & have & join & give & present here \\
8 & write & paint & remember & click here & live & spread \\
9 & tell & request & look & share & can read & test \\
10 & make & recommend & live & watch & get & apologize \\
11 & do & uncover & say & support & call & be have \\
12 & mean & search & take & crosspost & bring & conclude \\
13 & hear & 've make & will be & drink & be not & project \\
14 & get & maybe 'm & give & read here & make & will update \\
15 & be not & love most & believe & mix & go & must not allow \\
16 & do not know & 'd like add & get & sign here & tell & demonstrate \\
17 & ask & first write & do think & order here & believe & determine \\
18 & feel & ever run & understand & forward & take & search \\
19 & find & 'm use & ask & follow & become & propose \\
20 & call & scroll & read & pour & think & 're live \\
\bottomrule
\end{tabular}
\end{table}

\begin{figure}[ht!]
     \centering
     
     \includegraphics[width=\textwidth]{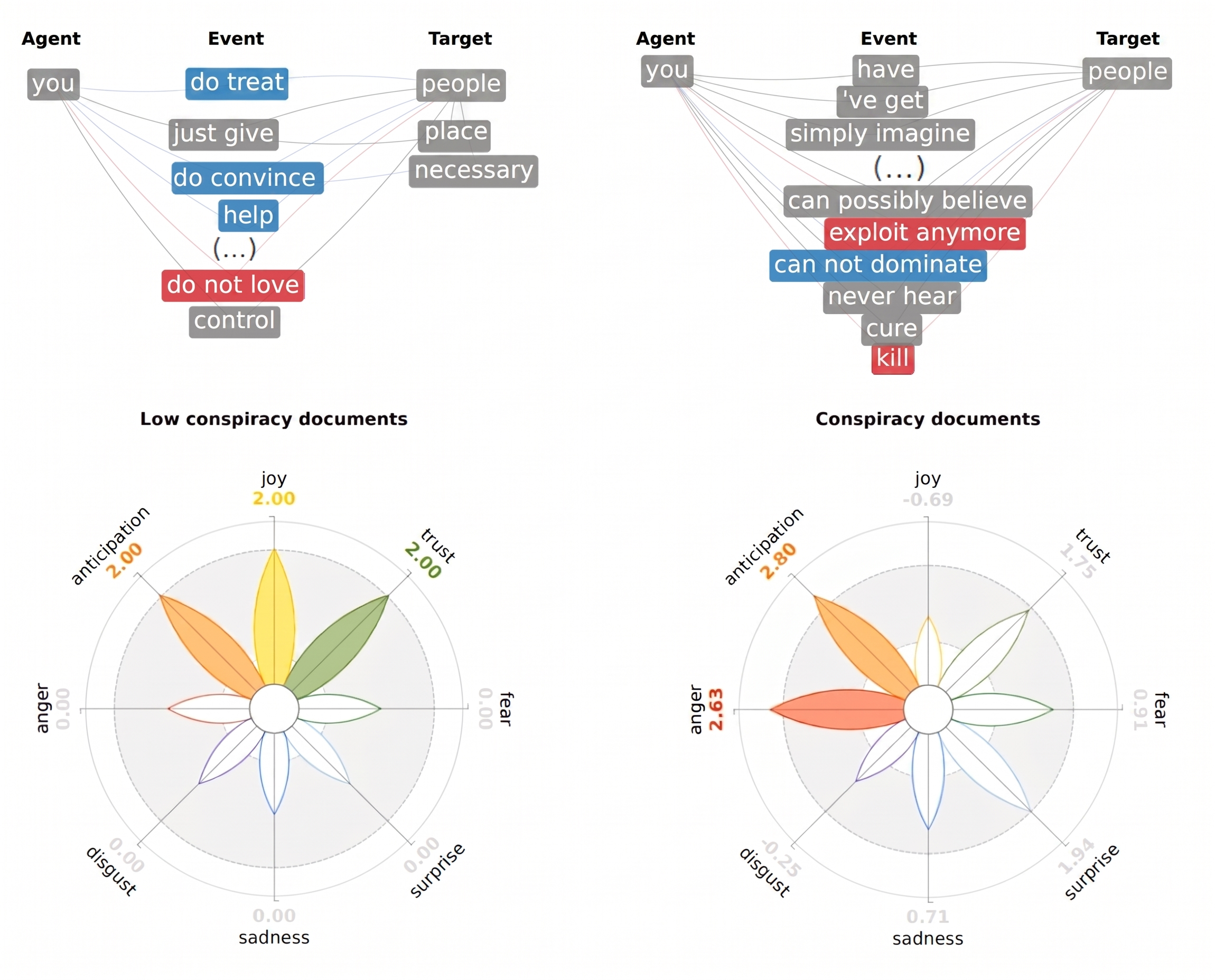}
     \caption{
     Analysis of ``You" acting on ``people" in the LOCO dataset. The top row displays SVO relationships extracted from low-similarity conspiracy documents (left) and high-conspiracy documents (right). The bottom row shows emotional flowers derived from the valence of verbs associated with these SVOs in the respective document sets.
     }\label{fig:LOCO_example}
 \end{figure}

\noindent \textit{\textbf{TEA Nets assess how narratives link actors and consequences with specific emotional actions.}}\\
\noindent Figure \ref{fig:LOCO_example} provides a comparative analysis of TEA Nets (top panels) and emotional profiles (bottom panels), differentiating between conspiracy (left panels) and low-similarity conspiracy documents (right panels).  The upper section displays TEA Nets, with the left panel representing a network derived from high-conspiracy documents and the right panel from low-similarity conspiracy documents. These networks are specifically filtered to highlight ``you" as a subject and ``people" as an object, showcasing a selection of verbs that form SVOs. The emotional flowers at the bottom of Figure \ref{fig:LOCO_example} reveal interesting differences. In low-similarity conspiracy documents (see Figure \ref{fig:LOCO_example}, bottom left), the actions used to bridge ``you" and ``people" report only weak positive emotions. 
Weakness comes from the fact that when EmoAtlas identifies only two actions providing the same emotion by default, a z-score close to the rejection region ($z\approx2$) is automatically reported by the library \citep{Semeraro_2025_EmoAtlas}. 

This means that in low-similarity conspiracy documents, narratives syntactically bridge ``you" and ``people" with slightly positive or mostly neutral jargon, like ``help" or ``treat", with exceptions including negative expressions like ``do not love". Instead, in high-conspiracy documents, actions show stronger anger components ($z\approx2.63$) and anticipation ($z\approx2.80$), which are further apart from the rejection region ($|z|<1.96$). This difference indicates that conspiracy narratives tend to express anger in actions specifying how the ``you" node relates to the ``people" node. This pattern is not mirrored in low-similarity conspiracy documents. Such a difference highlights how TEA Nets, when integrated with external tools like EmoAtlas \citep{Semeraro_2025_EmoAtlas}, can reveal nuanced differences in terms of specific portions of narratives across two groups.

\subsubsection{Case-Specific Discussion}
The TEA Nets framework, by explicitly identifying actors, actions, and consequences, allows the characterization of plot structures and roles within these narratives \citep{reiter_haas_2024_analysis_health_narratives}, overcoming traditional models' limitations in distinguishing actor from victim or attributing responsibility \citep{Sheikh_2023_causal_networks}.

Network-based linguistic measures highlight the practical utility of the TEA Nets Python library. Analysis of pronoun usage through degree, frequency, and Repetitiveness Index revealed self-referential and collective narrative patterns in high-conspiracy documents. This index combined frequency counts in texts with the syntactic network structure of Target-Event-Agent Networks, providing insights unavailable to mere frequency counts used in prior works \citep{Miani2022LOCO,klein2018topic,Fong_2021_language_conspiracy}. For instance, high Repetitiveness Index values for ``I" when considered as a subject in high-conspiracy texts reflected a self-referential linguistic pattern, consistent with prior research \citep{recordare_2024_conspiracy_theorists} and even with other linguistic markers, focused on ego-centered pronouns, found in the original paper introducing the LOCO dataset \citep{Miani2022LOCO}. 

The measurement of ``Agent-Event'' edge prominence quantifies associations between 
pronouns and verbs, revealing differences in verb usage across high-conspiracy and 
low-similarity conspiracy documents. High-conspiracy documents linked ``I'', ``you'', 
and ``we'' pronouns with verbs expressing beliefs and knowledge exchange, while 
low-similarity conspiracy documents showed a mixed pattern, including scientific terms 
but also commercial and website-navigation language; this suggests that boilerplate, 
advertisements, and source-specific content should be filtered before interpreting 
the contrast as a psychological difference. These patterns aligned with past 
social-media analyses \citep{Fong_2021_language_conspiracy,klein2018topic}. Earlier 
longitudinal work on Reddit showed that users who eventually migrate to 
\textit{r/conspiracy} already write more ``I'', ``you'', and ``we'' 
\citep{klein2018topic}. Cross-platform analyses on Twitter likewise report that both 
conspiracy influencers and their followers deploy these pronouns to construct in-group 
solidarity and recruit new adherents \citep{Fong_2021_language_conspiracy}. Our 
``Agent-Event'' edge prominence clarifies \emph{how} these pronouns are used to 
express actions: they are syntactically bound to epistemic verbs (e.g., ``I believe'', 
``we know'') expressing agency, beliefs and credibility, whereas low-similarity 
conspiracy documents foreground more impersonal science-process verbs (e.g., ``test'', 
``measure'', ``analyze''). This way of linking personal pronouns and epistemic verbs 
appears less frequently in low-similarity conspiracy documents. At the syntactic level, 
TEA Nets highlights a key organizational difference: high-conspiracy prose literally 
puts believing subjects at the center of the sentence, while low-similarity conspiracy 
prose suppresses personal agency in favor of impersonal scientific procedures.At the discourse level, large-scale graph analyses indicate that conspiracy corpora are more topically diverse yet tightly linked, producing texts that shift among themes while still maintaining internal coherence.\citep{Miani2022LOCO}.

Furthermore, TEA Nets uncovered a far higher prominence and repetitiveness of ``god" in high-conspiracy texts compared to low-similarity conspiracy narratives, indicating that mentioning this entity is a linguistic trademark of conspiratorial discourse, in alignment with past research showing that religion is a highly debated topic in high-conspiracy online texts on Reddit \citep{Fong_2021_language_conspiracy}. Past studies of social media datasets concerning conspiracy beliefs, also identified a negativity bias in the language adopted within conspiracy narratives \citep{Fong_2021_language_conspiracy,klein2018topic,Miani2022LOCO}. 
We followed this research direction and found the following trend, quantified via emotional profiling \citep{Semeraro_2025_EmoAtlas}: High-conspiracy actions linking ``you" and ``people" elicited stronger anger and anticipation compared to the actions expressed between the same actors and consequences but in low-similarity conspiracy texts. This specification not only aligns with past findings but also indicates quantitatively how Target-Event-Agent Networks can be used to test and quantify more specific patterns of language usage in corpora like conspiracy ones.

By providing structured syntactic and semantic information, the TEA Nets framework 
enables a more nuanced, context-aware quantitative assessment of emotional differences 
than traditional sentiment analysis \citep{hutto2014vader}, which often misses such 
distinctions due to its lack of contextual association with actors and actions. Future 
research could integrate edge prominence with narrative-frame detectors to test whether 
testimonial clauses like ``we-know'' or ``you-must-believe'' are used in analogous ways 
across various topics mentioned by high-conspiracy texts, such as crime, law, or power, 
as mentioned in \citep{Fong_2021_language_conspiracy}.

\subsection{Case 2: Agency in sexual-assault narratives on Reddit}\label{sect:assault}
The TEA Nets library offers a useful framework for information extraction from large corpora, particularly when the separation of active and passive verbal forms is crucial to the analysis. While traditional NLP methods that process surface-level linguistic data (e.g., dictionary-based and Bag-of-Words models) exhibit well-documented limitations with syntactic dependencies and the hierarchical organization of information \citep{Kenett_2022_Networks, Semeraro_2025_EmoAtlas}, the TEA Nets framework proves well suited to the study of sexual-assault datasets \citep{abramski2024voices} by extracting structured information that enables the identification of plot structures and the delineation of main roles, particularly the identification of the agent (i.e., who performed the action) in both active and passive sentences. We implemented a case study designed to illustrate this capability, employing a dedicated dataset of sexual assault subreddits comprising both posts and comments.

\subsubsection{Dataset}
To illustrate the utility of the TEA Nets framework, we present a case study using the dataset developed by \citep{abramski2024voices} for the study of agency in sexual assault narratives. This corpus comprises 6,418,677 posts and 3,846,797 comments from \textbf{\textit{r/sexualassault}}. Throughout this work, the term ``assault'' is used broadly to encompass the main topics discussed in this subreddit, including molestation, rape, assault, and abuse. For our analysis, we processed the Abramski dataset to produce two separate CSV files: one for posts and one for comments. In each file, the TEA Nets library was used to extract Subject-Verb-Object (SVO) triples from every sentence, including the identification of the agent in passive constructions. Since passive-voice resolution is not always certain, we relied on the \texttt{passive\_approx} binary flag defined in our methodology: a value of 0 indicates that the pipeline confidently identified an explicit agent, while a value of 1 indicates that uncertainty was detected (i.e., the grammatical patient was placed in the agent slot as an approximation). All approximated cases (\texttt{passive\_approx} = 1) were discarded, retaining only confident identifications for the analysis. In addition, each row was labeled for voice via the \texttt{is\_passive} flag, making it possible to focus the analysis on active or passive voice independently.

\subsubsection{Methods}
For each of the 10,265,474 rows in the initial dataset, \texttt{extract\_svos\_from\_text()} was used to extract SVOs and populate a structured DataFrame. We then removed empty texts, deleted duplicate rows, excluded rows for which no relevant assault-related Event was extracted, and retained only rows whose Event lemma matched one of four predefined lemmas: \textit{rape}, \textit{assault}, \textit{molest}, and \textit{abuse}. The number of rows retained after each filtering step is reported in Table \ref{tab:filtering_steps}.

\begin{table}[ht!]
\centering
\caption{Number of rows retained after each filtering step in the Reddit sexual-assault dataset.}
\label{tab:filtering_steps}
\begin{tabular}{@{}lc@{}}
\toprule
Filtering Step & Rows Retained \\
\midrule
Initial dataset & 10,265,474 \\
Removal of approximated passives and null agents & 2,270,075 \\
Selection of target lemmas (rape, assault, molest, abuse) & 315,173 \\
\bottomrule
\end{tabular}
\end{table}

Subsequently, using the tagged tabular data, we constructed a dedicated notebook for the analysis. The first post-processing step consisted of defining the set of \textit{Events} (verbs) relevant to this case study. We focused the analysis on four verbs: \textit{rape}, \textit{assault}, \textit{molest}, and \textit{abuse}, remapping each inflected form to its infinitive (e.g., \textit{raped} and \textit{raping} were both normalized to \textit{rape}). A second post-processing step was implemented to define and categorize the agents. Nouns were classified both by gender (Feminine, Masculine, Neutral) and by semantic category (Family, Spouse, Institutional, Vague, etc.). Each gender-by-category combination contains a set of nouns of interest as potential agents, such as \textit{guy}, \textit{man}, \textit{brother}, \textit{dad}, \textit{sister}, \textit{professor}, \textit{ex}, and others, derived through synonym extraction. We further filtered all rows to ensure only high-confidence cases were included (excluding \texttt{passive\_approx} = 1) and conducted separate analyses for active and passive voice forms. For this analysis, the posts and comments datasets were aggregated.

\subsubsection{Ethical considerations}
This case study was a secondary analysis conducted on the data gathered and fully anonymized originally by Abramski and colleagues \citep{abramski2024voices}, in accordance with institutional and platform-specific guidelines for research on public online data (GDPR EU-2016/679). We analyzed only material as present in the already anonymized dataset. We report here only aggregate statistics to impede any attempt of user identification or quote of individual posts. All derived outputs were inspected to minimize the risk of exposing personal information. No novel data was generated nor gathered.

\subsubsection{Results}
The aim of this case study was to identify the key agents perpetrating acts of sexual assault as described in the dataset. The first notable finding concerns the distribution of agent categories by gender and voice. Pronouns and vague descriptors, followed by generic and other categories, account for the vast majority of perpetrators in the dataset. The Family/Spouse category, however, represents a remarkably high share, one that warrants closer inspection in subsequent analyses. In terms of gender, Neutral agents constitute 72.3\% of the total, Masculine agents 22.8\%, and Feminine agents 4.9\%. Notably, terms classified as masculine occur approximately five times more often than terms classified as feminine among the retained agent terms. This should not be interpreted as a direct estimate of perpetrator gender, because many extracted agents are pronouns, vague references, or parser-derived approximations. Regarding voice, active forms account for 62.1\% of the analyzed sentences.

When all labels are aggregated to identify the most frequent linguistically identified agent terms associated with assault-related Events, the most common nouns are, in descending order: \textit{guy}, \textit{man}, \textit{brother}, \textit{dad}, \textit{people}, \textit{father}, \textit{cousin}. This ranking reveals a deeply troubling pattern in the dataset: after vague and generic terms, the Family/Spouse category constitutes the largest share of identified agents.

When the analysis is split by voice, the picture shifts in an even more alarming direction. Pronouns and vague terms, which dominate the active voice, are largely absent from the passive form, where the Family/Spouse category becomes the most frequent. Specifically, the most common agents in active constructions are \textit{I}, \textit{who}, \textit{he}, \textit{guy}, \textit{you}, \textit{she}, and \textit{it}; while in passive constructions, \textit{brother} appears as the most frequent, followed by \textit{someone}, \textit{family member}, \textit{man}, \textit{cousin}, \textit{friend}, \textit{father}, \textit{guy}, and \textit{dad}.

Table~\ref{tab:gender_voice_crosstab} reports the cross-tabulation between agent gender and grammatical voice.

\begin{table}[ht!]
\centering
\caption{Cross-tabulation of agent gender category by grammatical voice (active vs.\ passive) in the sexual-assault Reddit corpus.}
\label{tab:gender_voice_crosstab}
\begin{tabular}{@{}lccc@{}}
\toprule
Gender Category & Active & Passive & Total \\
\midrule
Feminine  & 6296 & 3824 & 10120 \\
Masculine & 28234 & 17069 & 45303 \\
Neutral   & 47431 & 29123 & 76554 \\
\midrule
Total     & 81961 & 50016 & 131977 \\
\bottomrule
\end{tabular}
\end{table}

\subsubsection{Case-Specific Discussion}
The results of this case study offer a sobering picture of how sexual violence is narrated and attributed in online survivor communities. Several findings deserve attention.

First, the prevalence of vague and pronominal agents in the active voice---terms such as \textit{he}, \textit{guy}, or \textit{who}---reflects a broader tendency in personal trauma narratives to describe perpetrators through diffuse or depersonalized language. This may reflect the psychological difficulty of naming the aggressor directly, or simply the conversational register of Reddit posts and comments, where shared context reduces the need for explicit identification \citep{o2018today,jaeger2014trauma}.

Second, and more strikingly, the passive voice reveals a markedly different distribution of agents, one in which Family/Spouse terms dominate. This is a linguistically and socially significant finding: when survivors adopt the passive construction, they are more likely to be describing assaults committed by family members \citep{bohner2001writing,o2018today}. This may suggest that passivization serves, consciously or not, as a distancing mechanism, a way to recount deeply intimate violations without foregrounding the identity of the perpetrator. The disproportionate presence of relational terms such as \textit{brother}, \textit{father}, \textit{cousin}, and \textit{family member} in passive constructions points to the compounded difficulty of narrating assault perpetrated within the family unit \citep{basile2022national}.

Third, the overall gender distribution of agents, with Masculine agents appearing nearly five times more frequently than Feminine ones, is consistent with established findings in the broader literature on sexual violence perpetration and reporting \citep{abramski2024voices,basile2022national,o2018today}.

Together, these findings demonstrate the analytical value of distinguishing active from passive voice when studying agency in trauma narratives. By leveraging the 
\texttt{is\_passive} and \texttt{passive\_approx} flags, active from passive 
constructions were separated and only high-confidence agent identifications were 
retained. 
Standard NLP approaches that flatten the syntactic structure would likely obscure precisely the patterns that are most meaningful here \citep{abramski2024voices,bohner2001writing}. The TEA Nets framework, by preserving and highlighting these distinctions, enables a more nuanced and linguistically grounded reading of how survivors represent assault and its perpetrators in online discourse.
At the same time, the results should be interpreted with caution, as many extracted agents are pronouns, vague references, or parser-derived approximations rather than definitive identifiers of perpetrators.

\subsection{Case 3: Comparing mental health transcripts of LLMs (CounseLLMe) and humans (HOPE)} \label{sect:CounseLLMe}
The TEA Nets framework introduces a new interpretable tool that is designed not only for the analysis of human-written texts, but also for texts produced by Large Language Models (LLMs). Uniquely, TEA Nets enables the automatic and explicit mapping of "Target-Event-Agent" relationships from extensive corpora, including mental health transcripts from both human and virtual (LLM-generated) patients. 
Thanks to its capability to capture non‑superficial linguistic features that go beyond word‑level counts or surface cues \citep{hutto2014vader,pennebaker2001linguistic}, TEA Nets represents a valuable framework to compare agents, actions and consequences as expressed in narratives by both humans and LLMs. Such a capability is highly relevant to the field of machine psychology, i.e. the emerging field where traditional psychological methodologies are leveraged to assess how closely LLMs can reproduce human behavior \citep{hagendorff2023machine}.

Evaluating the fidelity of an LLM-enhanced virtual patient is important \citep{wang2024patient, li2024leveraging}, especially when its language is compared with that of human patients \citep{deduro2025counsellme}. That is because, if a virtual patient based on an LLM can accurately mimic real patient behavior, LLMs could represent a valuable tool for training mental health professionals \citep{lozoya2025new, li2024leveraging}. In this context, we present an application of the Target-Event-Agent Networks framework to show its potential to uncover key distinctions in the language produced by humans and LLMs, specifically in the context of mental health.

\subsubsection{Dataset}
For this case study, involving the comparison between human and virtual patients in the mental health domain, we employed two pre-existing datasets: one comprising human transcripts and another consisting of synthetic content produced by LLMs. The first, Mental Health Counseling of Patients (HOPE; \citep{malhotra2022speaker}), is a dataset available for research purposes, including approximately 12,900 utterances produced by humans in counseling-like interactions, which were extracted from YouTube videos and subsequently transcribed. 
The second dataset, CounseLLMe \citep{deduro2025counsellme}, contains 400 synthetic counseling dialogues generated through interactions between two LLMs, with one model simulating a patient and the other acting as a therapist. This resource includes conversations generated by multiple models (OpenAI's GPT-3.5, Anthropic's Claude 3 Haiku, and LLaMAntino \citep{basile2023llamantino}) across both English and Italian. In our specific case, to match human data, we included only the English subset. 
CounseLLMe is publicly available at \url{osf.io/2ay8d}. 

\subsubsection{Methods}
The data pre-processing followed similar steps to those explained for the previous case study (Section \ref{sect:conspiracy_narrative}). 
We applied the function \texttt{teanets.extract\_svos\_from\_text()} to the combined utterances within each transcript, processing all utterances collectively rather than individually. This resulted in multiple SVOs extracted for each patient category ($N_{HOPE} = 212$, $N_{GPT}=100$, $N_{Haiku}=100$). Finally, individual patients' SVOs were concatenated. To prevent duplicate entries in the final \texttt{DataFrame}, the SVO IDs were manually incremented to ensure that identical dyads (S-V, V-O) did not share the same identifier. This process resulted in three distinct SVO \texttt{DataFrame}s representing human (HOPE dataset) and artificial patients (GPT and Haiku, from CounseLLMe) according to the TEA Nets framework. As in Section \ref{sect:conspiracy_narrative}, we combined TEA Nets with other approaches, such as EmoAtlas \citep{Semeraro_2025_EmoAtlas}, but additionally we employed a psycholinguistic dictionary including concreteness ratings for more than 40,000 lemmas \citep{brysbaert2014concreteness} and the NRC Emotion Intensity Lexicon \citep{mohammad2018intensities} to showcase the versatility of our framework.

\subsubsection{Results}
\begin{figure}[!htbp]
    \centering
    \includegraphics[width=0.85\linewidth, trim={0 20 0 20}, clip]{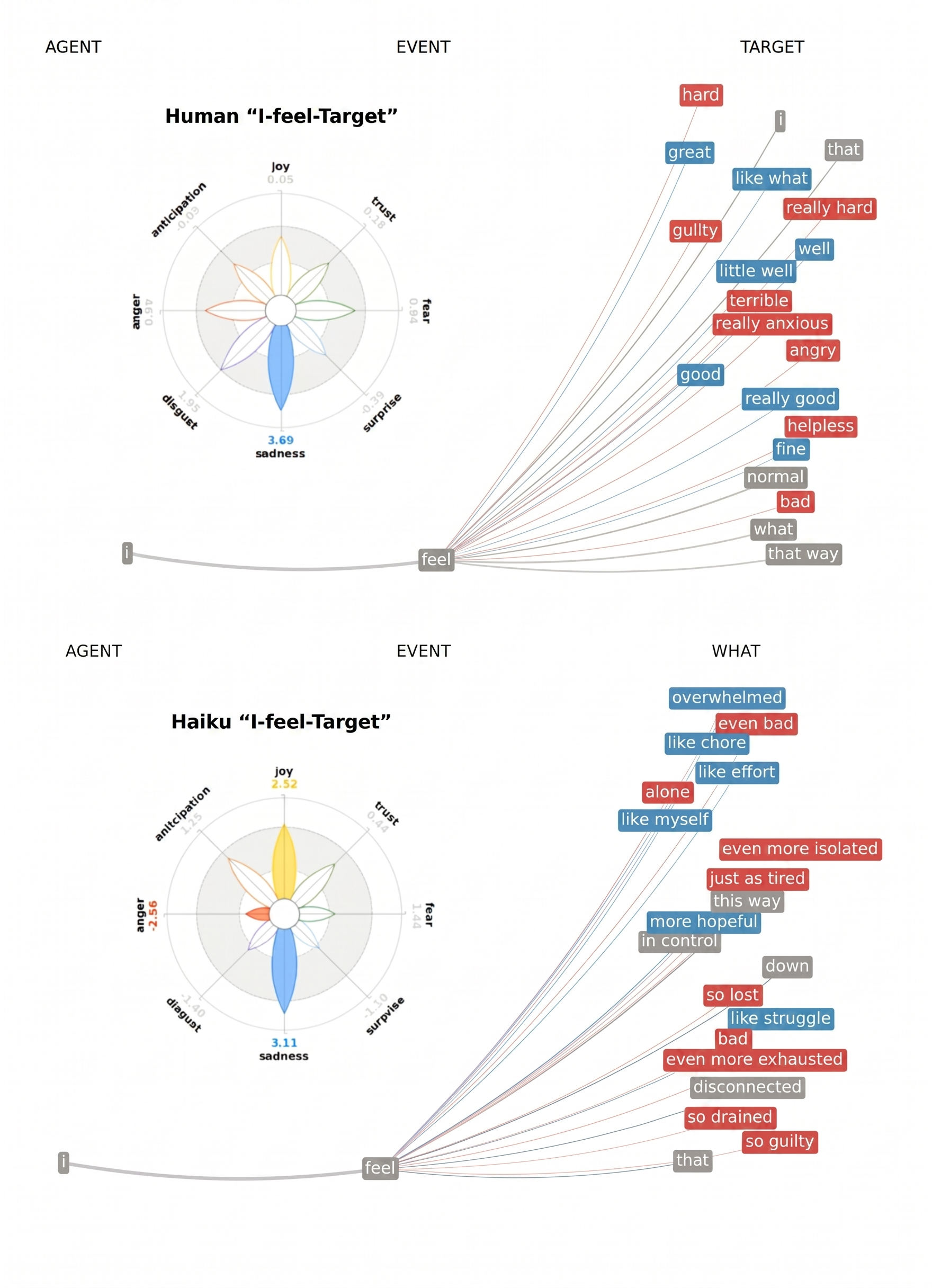}
    \caption{Top 20 most frequent objects (``Targets") associated with the ``I-feel" subject-verb pair in transcripts. Human patients are shown on top and Claude 3 Haiku simulated patients are shown on the bottom. These network vizualisations compare self-reported emotional states and experiences expressed through natural language across the two groups. On the top left of each plot, the emotional flower is generated using EmoAtlas, depicting the emotions expressed by the top 20 ``I-feel-Target" structures.}
    \label{fig:feel_top20_comparison}
\end{figure}

\textit{\textbf{LLMs and human patients express significant levels of sadness, but humans do so with greater intensity.}}
Empirical evidence shows that patients' willingness to share their emotions during psychotherapy sessions is strongly linked to better treatment outcomes \citep{peluso2018therapist}. To better understand how such disclosure is manifested in language, we turn our attention to a key linguistic marker of emotional expression, the verb ``feel", which in counseling transcripts can carry profound semantic significance and can signal a form of personal disclosure \citep{doell2013word}. To uncover what patients actually disclose, we adopt the TEA Nets framework to examine how the first‐person subject (``I") pairs with the verb ``feel" and identify the most frequent objects (``Targets") in that construction. 
In Figure \ref{fig:feel_top20_comparison} (top), we present the visualization of the 20 most prominent unique objects (``Targets") related to the dyad ``I-feel" (``Agent-Event") using \texttt{tea.plot\_svo\_graph()}.

Patients simulated by Claude 3 Haiku reveal a declarative and hyperbolic language related to ``feel". For example, the SVOs ``I-feel-even bad" or ``I-feel-so lost" highlight the negativity expressed by LLMs when impersonating clinical patients, prompted to manifest depressive symptomatology \citep{deduro2025counsellme}. Similarly, humans also use highly negative expressions (e.g., ``I-feel-unfulfilled", ``I-feel-drained"). 

To systematically assess the representation of specific emotions across the entire set of SVOs, we applied the EmoAtlas tool \citep{Semeraro_2025_EmoAtlas}. Notably, the results indicate the presence of significant levels of joy in Haiku's outputs when compared to a random baseline ($z \approx 2.32$), a pattern that was not observed in human or GPT-generated texts (see Figure \ref{fig:fig_flower_gpt}, Appendix \ref{apx:GPT}). Furthermore, expressions of anger are markedly less frequent in Haiku's output than would be expected by chance ($z \approx -2.56$). This prevalence of joy-oriented and scarce anger-related expressions in the "I-feel-Target" triad may reflect inherent properties of Haiku's training data, which prioritize accommodation and the maintenance of a positive conversational tone even when the system is prompted to emulate individuals experiencing depressive symptoms.

Separately, higher-than-random levels of sadness are found in humans ($z \approx 3.69$), Haiku ($z \approx 3.11$) and GPT ($z \approx 4.54$). Here, human and AI-generated texts successfully captured the core emotional markers associated with a clinical condition such as depression \citep{deduro2025counsellme}. 
To further clarify whether these differences reflect not just the prevalence but also the intensity of emotional expression, we conducted a subsequent analysis using the NRC Emotion Intensity Lexicon (NRC-EIL; \citep{mohammad2018intensities}) on the whole set of ``I"-``feel"-``Target" structures. Focusing solely on words expressing sadness, we compared the emotional intensity of sadness-laden language between Haiku and humans. Interestingly, despite the greater frequency of sadness observed in Haiku's language (among the top 20 SVOs), human-produced expressions of sadness tended to use words with significantly higher intensity ($U = 1243.5, p = .036$). This pattern was also present for expressions of disgust, where humans again relied on higher-intensity language ($U = 242.5, p = .042$). However, when analyzing the intensity of words associated with other emotions, we did not observe significant differences between groups.

These findings align with the recent literature documenting LLMs' tendency to generate emotionally moderate responses rather than extreme affective states \citep{fazzi2025don}. Similarly, LLMs were shown to fail in accurately mirroring emotional expressions, even if equally showcasing significant levels of sadness, present in human language \citep{huang2024apathetic}. Although this can provide a more engaging experience for the user in some applications, it raises issues in clinical training contexts where virtual patients are employed, especially if they fail to mirror the intensity and variability of human emotional expression. Prior evidence suggests that increasing the realism of simulated interactions improves training effectiveness for clinicians \citep{garcia2024enhancing}, highlighting the need for LLMs to more accurately model the variability of human emotional expression in such settings.

\begin{figure}[!htbp]
    \centering
    \begin{overpic}[width=1\linewidth]{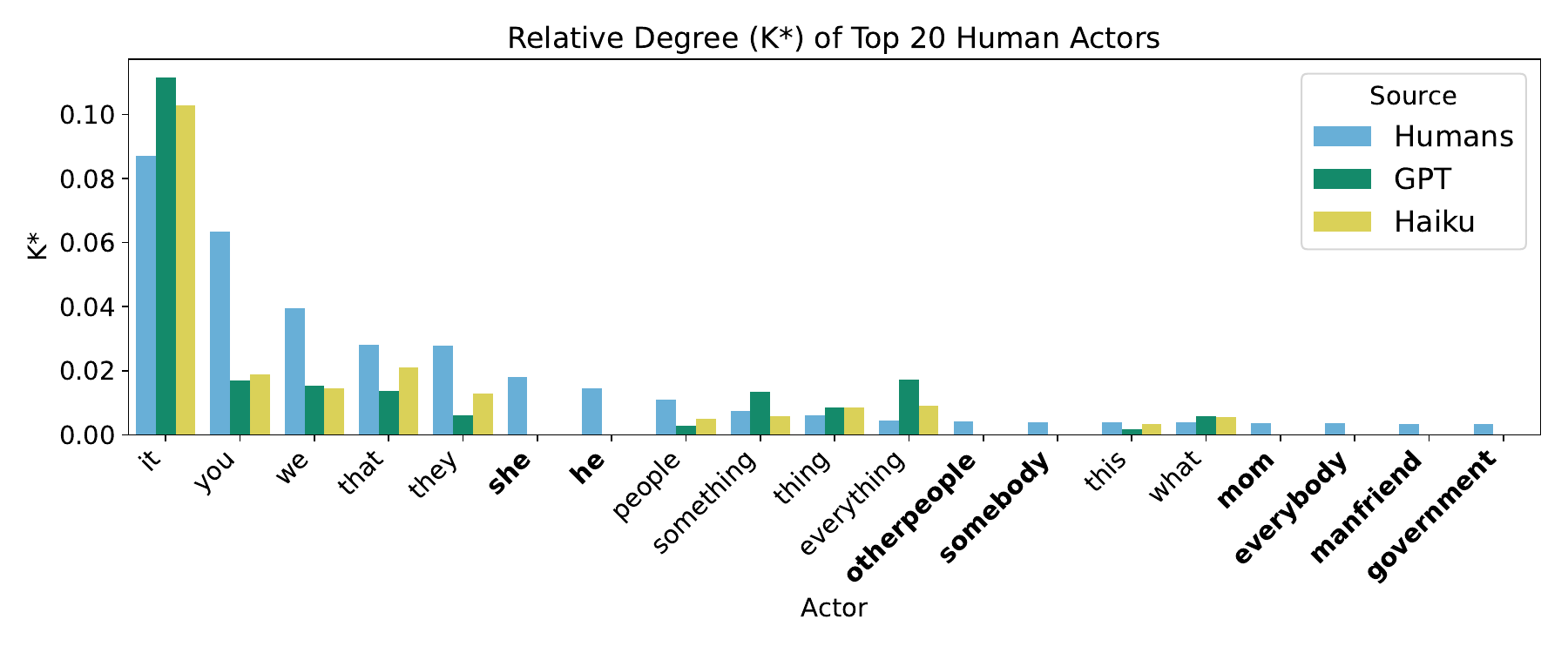}
        \put(2.85,2.85){\color{white}\rule{10pt}{170pt}}
    \end{overpic}
    \caption{Relative degree ($K^*$; computed within each group) for the top 20 most central actors found in human transcripts. The corresponding $K^*$ values for the same set of actors are also provided for both LLM-generated (GPT and Haiku) TEA Networks. For clarity, the pronoun ``I'' is excluded, as it has the highest relative degree across the three groups: humans ($K^* = .675$), GPT ($K^* = .639$) and Haiku ($K^*=.835$). Actors unique to human transcripts are shown in bold.}
    \label{fig:top20_centrality}
\end{figure}

\textit{\textbf{LLMs use significantly less-concrete actors (``Agent") when compared to humans.}}
Linguistic analyses of counseling sessions have revealed that pronoun choice can both signal symptom severity and track therapeutic progress \citep{van2004changes}.  For instance, research has shown that the use of first-person pronouns correlates with depressive symptoms ($r = 0.13, \alpha = 0.05$; \citep{edwards2017meta}). 
To examine this effect, we analyzed the centrality of actors (subjects) in human and LLM-generated data. Figure \ref{fig:top20_centrality} presents the 20 most central actors for humans (HOPE), as measured by the within-class relative degree ($K^*$, see Section \ref{sec:measures}), along with the values for the same actors found in LLMs' synthetic data (CounseLLMe).

Across the three datasets, the first-person singular pronoun ``I" has the highest relative degree, underlining the self-referential language expected from patients in a counseling session \citep{rogers2012client}, a pattern that extends to LLM-generated dialogues \citep{deduro2025counsellme}. In addition, the pronoun ``we", when used as a subject, displays non-negligible centrality across all sources: humans ($K=381$, $K^*=.0279$), GPT-3.5 ($K=50$, $K^*=.0154$), and Claude 3's Haiku ($K=70$, $K^*=.0144$). These results may suggest the presence of therapeutic alignment in human interaction, as previously shown in humans by \citet{nikzad2025evolution}, and potentially indicate that a similar process can emerge in a simulated counseling scenario generated by LLMs.

Additionally, humans produce a wider set of ``Agents" that are not reflected in LLMs. Human speakers, in addition to situating themselves in their narratives, sometimes refer to other individuals (e.g., ``she", ``he", ``mom") and broader social contexts (e.g., ``government"), thereby embedding their experiences within specific autobiographical and interpersonal narratives \citep{conway2000construction}. LLMs instead tend to use vaguer language. To quantitatively assess this, we assign a concreteness score to each unique ``Agent" extracted within TEA Nets, for the three classes using a large psycholinguistic dataset containing 40,000 lemmas associated with concreteness scores \citep{brysbaert2014concreteness} (see Appendix~\ref{apx:tab_who_counsellme} for details). We did not find significant differences between LLMs ($U_{GPT-Haiku} = 104,292$, $p = .507$). However, concreteness scores in subject words were significantly different between humans and LLMs ($U_{Humans-GPT} = 242,918$, $U_{Humans-Haiku} = 247,203$, $p < .001$). Human patients produced responses with a higher median concreteness score ($Median_{Humans} = 3.20$) compared to both GPT ($Median_{GPT} = 2.76$) and Haiku ($Median_{Haiku} = 2.70$) simulated patients. Furthermore, the inter-quartile range ($IQR$) of the distributions produced by the LLMs was notably narrower ($IQR_{Haiku} = 0.93$, $IQR_{GPT} = 0.93$) compared to that observed in the human data ($IQR_{Humans} = 1.46$).

These findings are consistent with recent studies showing that LLMs tend to produce generic language that lacks grounding in real-world scenarios \citep{birhane2024large, pavlick2023symbols} and exhibit reduced semantic richness compared to human language \citep{guo2024curious}.

\subsubsection{Case-Specific Discussion}
Using the TEA Nets framework, we introduce a novel methodology for comparing human and Large Language Model (LLM) responses in the context of machine psychology \citep{hagendorff2023machine}. In the first analysis, we isolated objects (Targets) associated with "I-feel" networks in both humans and LLMs, comparing 200 texts generated by LLM-simulated clients with 212 counseling-like responses produced by human patients.

Using EmoAtlas \citep{Semeraro_2025_EmoAtlas}, we found that both humans and LLMs express significant levels of sadness within "I-feel-Target" sentences, though LLMs do so with lower intensity, as measured by NRC-EIL \citep{mohammad2018intensities}, possibly due to safeguarding mechanisms in AI training that lead to the suppression of extreme expressions of negative emotions \citep{fazzi2025don}. 
Interestingly, Claude 3's Haiku showed significant levels of joy, even when simulating a patient with depression, highlighting discrepancies with humans. In the second analysis, we adopted the TEA Nets library to identify actors mentioned in the narratives produced by both humans and LLMs. We subsequently applied the norms from \citet{brysbaert2014concreteness} to compare how concrete these actors were in human and LLM-generated texts and found that models like Claude 3's Haiku and GPT-3.5 demonstrated a preference for less concrete actors and exhibited less variability in 
this psycholinguistic dimension compared to humans, suggesting a reduced grounding in real-world scenarios \citep{guo2024curious, birhane2024large}.. 

Overall, these discrepancies in emotional intensity and psycholinguistic features, highlight the current challenges in achieving realistic LLM-enhanced virtual patients \citep{garcia2024enhancing}. Nevertheless, when emotional intensity is disregarded, we found alignment in terms of sadness from both humans and LLMs. We believe this alignment underscores the potential for further refinement in AI for mental health. For this reason, we stress the importance of accurate and psychologically informed prompt engineering \citep{chen2023llm} or fine-tuning \citep{wang2024patient} to better capture authentic human experiences in future virtual patient simulations. Nevertheless, our results indicating similar expression of emotions in "I-feel-Target" networks (although often failing to capture their full intensity) suggest some promise for developing AI clients in mental health training. A fine-tuned version of GPT-4 has recently been shown to improve virtual patients' effectiveness for training purposes compared to baseline models \citep{wang2024patient}. At the same time, models' role-playing capabilities are improving \citep{wang2025characterbox}, and new approaches using psychologically-informed prompt engineering have helped better mimic the complexity of real-world scenarios \citep{lozoya2025new}. 

Overall, our findings suggest that the LLM-simulated patients examined here still diverge from human counseling transcripts in emotional intensity and concreteness of actors. These differences should be addressed before such simulations are used as high-fidelity substitutes for human patient interactions in psychotherapy training. TEA Nets revealed marked differences in both emotional intensity and real-world grounding between humans and LLMs, which may reduce the effectiveness of using AI systems in training scenarios. Increasing the realism of LLM-enhanced virtual patients is critical, as immersive role-playing and high-fidelity simulations are known to enhance learning transfer to real clinical settings \citep{garcia2024enhancing, reger2021virtual}. We argue that TEA Nets are particularly valuable for informing prompt engineering for virtual patients. By extracting structured syntactic and semantic information from both human and LLM-generated texts, Target-Event-Agent Networks make it possible to explicitly identify divergences between LLM responses and authentic human conversations. These insights can guide the refinement of prompts to support more realistic and clinically useful future virtual patient interactions.


\section{Discussion} \label{sect:discussion}
Narratives are central to human cognition, shaping self-understanding, interpersonal perceptions, and event comprehension \citep{sarbin1986narrative, mcadams2011narrative}. This work introduces the TEA Nets framework as a powerful and novel methodological approach to analyze narratives. A Target-Event-Agent Network can provide an interpretable approximation of how narratives assign agency, actions, and consequences \citep{reiter_haas_2024_analysis_health_narratives}, enabling researchers to study how people assign responsibility and perceive causal relationships \citep{Sheikh_2023_causal_networks}. 

Whereas traditional methods of Natural Language Processing, such as Bag-of-Words, focus on word frequency or co-occurrence, TEA Nets leverage both syntactic and semantic information to automatically construct networks capturing three central narrative components: subjects (Agents), verbs (Events), and objects (Targets). This can be valuable in behavioral and text analysis research. For instance, in a text expressing distress \citep{deduro2025counsellme}, understanding the perceived agent of suffering, the harmful action, and the affected entity can provide specific insights into the user's psychological state and narrative framing \citep{abramski2024voices}. TEA Nets could also be useful for tracking therapeutic progress \citep{nikzad2025evolution}. In the context of Obsessive-Compulsive Disorder (OCD) treatment, a client's shift from "I washed my hands" to "The OCD made me wash my hands" demonstrates externalization (i.e., viewing symptoms or problems as separate from the self), a key therapeutic goal \citep{barrett2008evidence} that TEA Nets can systematically detect and quantify by specifically extracting the subject (and not the object) from transcripts.

The Target-Event-Agent Networks framework uniquely captures actors, actions and consequences from large corpora. By doing so, it helps identify semantic and psychological implications within text by providing a clear, structured representation of relationships between key elements in the narrative. \citet{Teixeira_2021_structure_notes} showed that personal pronouns like ``I" play central roles in suicide notes. In this context, the TEA Nets framework presents an opportunity to extend this analysis by enabling the identification of ``I" as subject or object, and integrating emotional valence with its associated actions. Distinguishing the syntactic role of concepts expressed in narratives can be crucial. In psycholinguistics, the grammatical role of concepts (e.g., subject versus object) has been shown to be more influential than mere mention order or salience in pronoun resolution \citep{hert2024importance}. Furthermore, \citet{abramski2024voices} provided a strong example of why mapping actors, actions, and consequences can be relevant in psychological inquiries. Consider the following two sentences: (i) ``I was raped by my uncle" and (ii) ``I was raped, while my uncle was watching". In (i) ``uncle" is the perpetrator, while in (ii) ``uncle" is a bystander. Traditional cognitive network science approaches do not account for the different roles attributed to ``uncle".  In contrast, by systematically extracting actors, actions, and consequences, TEA Nets make it possible to determine whether "uncle" is linked directly with the action of sexual assault or not, distinguishing between the two cases. Such distinctions allow for the extraction of role-specific semantic frames and better understanding of the emotions tied to "uncle" in each context. This nuanced differentiation is crucial, as the role of bystanders and perpetrators can have distinct psychological impacts in cases of sexual assault \citep{labhardt2025intervene}.

In order to carry out analyses like the one described above, we have made TEA Nets accessible as a Python package (\url{https://github.com/MassimoStel/TEA_Networks}). In the following, we discuss TEA Nets in terms of interpretability, case studies, relevant cognitive-science literature, and limitations.

\textbf{TEA Nets support interpretable text analysis.} A core strength of our framework lies in its interpretability. Such interpretability is achieved on two levels: first, through a transparent and explainable process of network construction, and second, by enabling clear, quantitative, and psychologically meaningful analyses of the resulting networks. In the network extraction process, we rely on the AI's ability to perform syntactic parsing (using spaCy; \citep{spacy_website}), an important component in the overall process of creating Target-Event-Agent Networks. In this way, artificial intelligence is used to detect structural features of language rather than to work directly with vector embeddings. Thanks to this difference, TEA Nets provide a more interpretable framework than purely AI-driven approaches for extracting agents, actions, and consequences from unstructured textual data \citep{cheng2024potential}. In addition, TEA Nets adopt the VADER sentiment lexicon \citep{hutto2014vader} to label key elements in narratives with sentiment valence information. By doing so, TEA Nets introduce an affective dimension to the network representation while preserving transparency using a dictionary-based approach \citep{feuerriegel2025using}. The structured output of the Target-Event-Agent Networks framework enables network-based analysis of texts using measures like degree centrality, allowing researchers to gain quantitative, interpretable, and psychologically meaningful insights from narratives \citep{Siew_2019_CogNetSci, abramski2024voices, stella2020text}.

Furthermore, TEA Nets support systematic comparisons of linguistic patterns across different categories of texts, allowing researchers to detect and examine subtle yet significant variations in language use. This is particularly important in several psychological fields, including the sensitive domain of mental health \citep{de2023benefits}, where language produced by humans can be compared against that of LLMs to gain insights into the limitations and biases of current models.

\subsubsection*{Limitations}
Despite their advantages, Target-Event-Agent Networks have limitations.
First, TEA Nets are naturally prone to ambiguities that cannot be entirely resolved from text. For instance, our current TEA Nets implementation relies by default on FastCoref for coreference resolution \citep{Otmazgin2022FcorefFA}. We chose FastCoref for its lightweight nature; however, coreference resolution remains an intrinsically challenging task in computational linguistics. Even advanced solvers like FastCoref and Stanza (when used as an alternative; \citep{qi2020stanza}) do not consistently achieve high accuracy scores across all contexts. It is important to note, however, that the modular architecture of the TEA Nets Python implementation allows for future upgrades: as improved coreference resolution techniques are developed, they can be readily integrated into the library. A second limitation arises when applying TEA Nets to data extracted from spoken interactions, such as therapist transcripts. This speech-to-text process necessarily leads to the loss of prosodic features such as tone, pitch, and rhythm, which may convey crucial linguistic information. Consequently, text-based analyses are inherently limited in their capacity to capture elements such as sarcasm or emotional nuance \citep{Semeraro_2025_EmoAtlas}, which often rely on intonation and other non-textual cues. Researchers should therefore interpret findings in light of this limitation, emphasizing the semantic and syntactic content that is reliably accessible through text analysis. The third limitation, also shared with textual forma mentis networks \citep{stella2022cognitive}, stems from the use of a dictionary-based lexicon to annotate the nodes with negative, neutral, or positive valence. Here, we chose VADER \citep{hutto2014vader} for sentiment analysis because of its transparency and generally strong performance. However, words mentioned in a specific context might undergo shifts in their meaning (e.g., "sick" as a negative descriptor in everyday speech versus a positive slang term), potentionally leading to misclassified sentiment. For this reason, researchers are encouraged to interpret concepts within the broader framework of their relational networks, rather than considering them in isolation, to avoid drawing misleading conclusions. A further limitation is that Subject–Verb–Object units mostly correspond to individual concepts or periphrases. As a result, the system may struggle to accurately represent more complex sentence structures, particularly nested or hierarchical clauses. This limitation is partly mitigated by the use of hypergraphs \citep{citraro2023hypergraph}, so that individual nodes in networks can be whole sentences or subgraphs rather than isolated words. Last but not least, TEA Nets currently supports only English. Since Scandinavian languages share the same SVO structure of English \citep{holmberg1998word}, TEA Nets could be potentially expanded to other non-English languages while being careful about passive form construction. However, this extension depends on the availability of future libraries supporting SVO detection in non-English languages.

\section{Conclusion}
In this work, we introduced the TEA Nets framework as an innovative computational tool, grounded in cognitive network science, Natural Language Processing, and Artificial Intelligence. TEA Nets are specifically designed to extract and represent actors, actions and consequences from textual data. We implemented this framework as an open-source Python library to support its practical use in psychological and multidisciplinary research settings. We provide a comprehensive description of the processing pipeline, detailing how unstructured text inputs are systematically transformed into structured TEA Nets representations. We showcase three practical implementations of the TEA Nets Python library across conspiracy narratives, online sexual-assault narratives, and counseling transcripts involving human and LLM-simulated patients.
These applications demonstrate that the TEA Nets framework offers a granular, interpretable, and powerful approach for analyzing relational dynamics in language. By complementing existing text analysis methods, the TEA Nets framework opens new avenues for research in behavioral and social sciences.

\section{Acknowledgments}

This work was supported by the contribution of the Ministero Dell'Università e della Ricerca (MUR) according to Decreto N. 23178 of 10 dicembre 2024 - Bando FIS 2. The authors acknowledge support from CALCOLO, funded by Fondazione VRT, for support with the computational infrastructure simulating LLMs. We acknowledge Salvatore Citraro for valuable feedback on a preliminary version of this manuscript.

\bibliography{references}

\appendix

\renewcommand{\thefigure}{S\arabic{figure}}
\setcounter{figure}{0}

\section{Supplementary Information}

\subsection{Python Library: \texttt{teanets}}\label{sect:python_library}

The TEA conceptual framework is implemented as an open-source Python library, publicly available at \url{https://github.com/MassimoStel/TEA_Networks} (Accessed 28/04/2026): \texttt{teanets}. This library provides a suite of computational tools that enable researchers to apply the TEA methodology to textual analysis.  
The library employs a modular design that separates the analysis pipeline into distinct, function-specific components. These key components, detailed below, enable  textual analysis within the framework: (i) \texttt{textloader}; (ii) \texttt{nlp\_utils}; (iii) \texttt{svo\_extraction}; (iv) \texttt{svo\_to\_graph} and network analysis; (v) \texttt{teanets.analytics}; (vi) \texttt{resources}.

The \texttt{textloader} module handles text cleaning and coreferencing. The library supports integration with the \texttt{FastCoref} coreference solver \citep{Otmazgin2022FcorefFA}  (or alternatively, with \texttt{Stanza}; \citep{qi2020stanza}).

The \texttt{nlp\_utils.py} module provides NLP utilities, including text processing via spaCy's NLP pipeline, accessible through the \texttt{spacynlp} function. This module also incorporates sentiment analysis. The \texttt{compute\_valence} function, using the VADER module \citep{hutto2014vader}, measures the emotional valence of textual data at the level of each node. The resulting valence scores can be used to annotate graph nodes with sentiment labels.

The \texttt{svo\_extraction.py} module is responsible for extracting Subject-Verb-Objects triples (SVOs) from texts that have been syntactically parsed with \texttt{spaCy}.  
Using helper functions such as\texttt{get\_verb\_subjects}, \texttt{get\_verb\_objects}, and \texttt{get\_verb\_phrase} this module identifies the key components of SVOs. Furthermore, \texttt{svo\_extraction.py} identifies semantic relationships by checking for synonymy between words within the same syntactic roles (layers), using WordNet \citep{miller1995wordnet} via the \texttt{are\_synonymous} function.
 
Beyond data extraction, the library offers tools for analysis and visualization. It enables the construction of \texttt{NetworkX} graphs \citep{Aric2008NetworkX} from SVOs via the \texttt{svo\_to\_graph} function. Nodes are typed according to their syntactic role: subject, verb, or object, corresponding to the TEA roles Agent, Event, and Target. Edges within the network can represent both syntactic and semantic relationships. The \texttt{svo\_to\_graph} function includes options for filtering and sentiment-based colour-coding. Thanks to \texttt{filter\_svo\_dataframe\_by\_tea}, it is possible to filter the SVOs based on the TEA role. Users can also calculate weighted degree centrality for specific roles via \texttt{tea\_weighted\_degree\_centrality}, and obtain an overview of central nodes with \texttt{tea\_degree\_centrality\_overview}.
 
The \texttt{teanets.analytics} module provides comprehensive tools for manipulation and export of TEAs.  Among its utility functions are \texttt{merge\_svo\_dataframes}, which facilitates the merging of multiple DataFrame objects, and \texttt{export\_hypergraphs}, which enables the export of syntactic hypergraphs. In addition, the module provides functionality for saving data in CSV format.
 
Finally, the \texttt{resources} module houses essential internal resources, including various resource files utilized internally by the pipeline, such as specific word lists used for coreference resolution or for determining sentiment valence. These resources support the internal operations of the pipeline and facilitate the implementation of key functionalities.

\subsection{Guide for the TEA code implementation}\label{sect:code_tutorial}

This section provides a step-by-step guide to using the \texttt{teanets} Python library in a Google Colab notebook for extracting SVOs and visualizing TEAs. Readers can try it directly \href{https://github.com/MassimoStel/TEA_Networks/blob/main/Docs%20%26%20Guides/TEA_Paper_tutorial.ipynb}{here}.

\begin{figure}[ht!]
    \centering
    \includegraphics[width=1\linewidth]{figures/user_tutorial.png}
    \caption{
    Snippet of the Python code demonstrating the process of SVO triplet extraction and TEA generation from a sample text using the \texttt{teanets} library. The output showcases the tabular SVO \texttt{DataFrame}.
    }
    \label{fig:user_tutorial}
\end{figure}

Figure \ref{fig:user_tutorial} displays a minimal working example of using the \texttt{teanets} package, following four simple steps:

\begin{enumerate}
    \item \textbf{Install the Target-Event-Agent Networks Library.} The first command of the code snippet installs the library by running \texttt{!pip install -q git+https://github.com/MassimoStel/TEA\_Networks}. This command executes a shell command in the Colab environment, installing the \texttt{teanets} library directly from its GitHub repository.
    \item \textbf{Load Necessary Modules.} The following two commands (\texttt{import teanets as tea} and \texttt{import teanets.analytics}) import the main \texttt{teanets} library and the \texttt{analytics} submodule, respectively. These imports make the functions defined within the modules accessible for use in the notebook, and they ensure that all functions and methods defined within these modules are available for use throughout the current notebook.
    \item \textbf{Extract SVOs from Text.}  The next command defines a text string, assigned to the \texttt{text} variable (see Figure \ref{fig:infograph_introd}A). The \texttt{tea.extract\_svos\_from\_text(text)} function processes this text to extract SVOs that encode the TEA Nets relationships within the sentences. The \texttt{display(svo)} command outputs the extracted SVOs as a pandas \texttt{DataFrame} (see Figure \ref{fig:infograph_introd}D). 
    \item \textbf{Plot the TEA Network.} The final command (\texttt{tea.plot\_svo\_graph(svo, custom\_font = 20)}) takes the extracted \texttt{svo} \texttt{DataFrame} and plots it as a Target-Event-Agent Network graph, as presented in Figure \ref{fig:infograph_introd}E. The \texttt{custom\_font = 20} argument adjusts the font size for improved readability of the node labels in the generated plot.
\end{enumerate}

\subsubsection{Supplementary and Advanced Guides}

To support researchers in applying and verifying the TEA framework, the \href{https://github.com/MassimoStel/TEA_Networks/tree/11a700a3309397866d2a894b531852e61443b39d/Docs%20%26%20Guides}{\texttt{Docs \& Guides}} folder contains the complete suite of interactive notebooks:
\begin{itemize}
    \item[--] \textbf{Starting Guide} (\href{https://github.com/MassimoStel/TEA_Networks/blob/main/Docs\%20\%26\%20Guides/Starting\%20Guide.ipynb}{\texttt{Starting\_Guide.ipynb}}): An extended overview of the pipeline architecture and configuration.
    
    \item[--] \textbf{Analyzing TEA} (\href{https://github.com/MassimoStel/TEA_Networks/blob/main/Docs\%20\%26\%20Guides/Analyzing\%20TEA.ipynb}{\texttt{Analyzing\_TEA.ipynb}}): Focuses on computing network metrics and analyzing the emotional valence of nodes.
    
    \item[--] \textbf{Actor Categorization} (\href{https://github.com/MassimoStel/TEA_Networks/blob/main/Docs\%20\%26\%20Guides/Actor\%20Categorization\%20Tutorial.ipynb}{\texttt{Actor\_Categorization\_Tutorial.ipynb}}): Shows how to group semantic agents into broader thematic categories for high-level analysis.

    \item[--] \textbf{Generating TEA Figures} (\href{https://github.com/MassimoStel/TEA_Networks/blob/main/Docs\%20\%26\%20Guides/Generating_TEA_Figures.ipynb}{\texttt{Generating\_TEA\_Figures.ipynb}}): Provides
  a step-by-step tutorial on how to generate and export the Target-Event-Agent network visual representations used in the repository documentation.

\end{itemize}

\subsection{GPT's "I-feel-Target" structures and emotional flower}\label{apx:GPT}

\begin{figure}[H]
    \centering
    \includegraphics[width=1
    \linewidth]{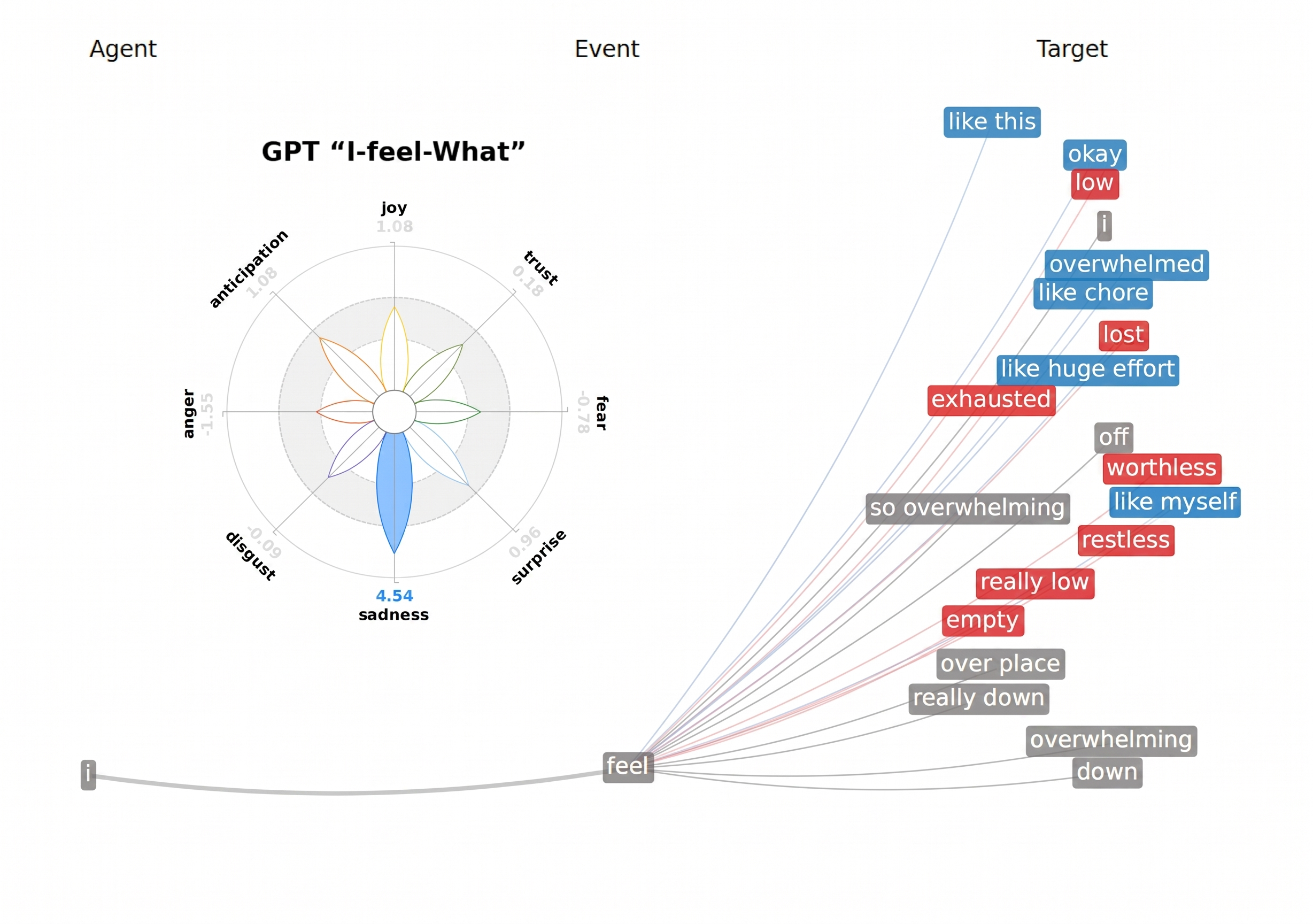}
    \caption{{Top 20 most frequent objects (``Targets") associated with the ``I-feel" subject-verb pair in GPT-3.5 transcripts.}}
    \label{fig:fig_flower_gpt}
\end{figure}

\subsection{Actor concreteness} \label{apx:tab_who_counsellme}
To compute the concreteness scores for each unique actor extracted from TEA, we employed the word concreteness norms developed by \citet{brysbaert2014concreteness}, which provide concreteness ratings for a large set of English words. We create one list entry for each word (e.g., ["big argument"] becomes ["big"], ["argument"]). We then take the set of actors from this list and assign a concreteness score to each element of the set using the dictionary form \citet{brysbaert2014concreteness}. Finally, we compared the resulting concreteness score distributions using the Mann–Whitney U test implemented in \texttt{scipy.stats.mannwhitneyu} in Python.

\subsection{Target-Event-Agent edges in conspiracy narratives}\label{apx:loco_edges}
Table \ref{tab:edges_freq} presents the frequencies of the top 20 ``Agent-Event" and ``Event-Target" edges for both high-conspiracy and low-similarity conspiracy documents. All frequency values are presented in thousands. 

\begin{table}[ht!]
\centering
\caption{Edge Frequency ($EdF$) of the top 20 ``Agent-Event" and ``Event-Target" edges in high- and low-similarity conspiracy documents. Values are expressed in thousands.}
\label{tab:edges_freq}
\footnotesize
\begin{tabular}{@{}llrllrllrllr@{}}
\toprule
\multicolumn{6}{c}{Conspiracy} & \multicolumn{6}{c}{Low-similarity Conspiracy} \\ \midrule
Who & Did & \multicolumn{1}{l}{EdF} & Did & What & \multicolumn{1}{l}{EdF} & Who & Did & \multicolumn{1}{l}{EdF} & Did & What & \multicolumn{1}{l}{EdF} \\ \midrule
it & be & 12.85 & be & what & 3.27 & it & be & 1.11 & be & what & 0.22 \\
that & be & 6.89 & do & what & 1.75 & this & be & 0.48 & be & likely & 0.22 \\
this & be & 5.75 & be & able & 1.43 & that & be & 0.45 & be & able & 0.20 \\
I & be & 4.77 & circulate\footnotemark[1] & today & 1.35 & you & see & 0.20 & be back & free shipping & 0.18 \\
we & be & 3.75 & circulate\footnotemark[1] & in kremlin & 1.35 & you & be & 0.20 & be back & at \% & 0.18 \\
you & be & 3.29 & be & part & 1.30 & we & be & 0.17 & be back & with\footnotemark[2] & 0.18 \\
I & think & 2.32 & be & true & 1.16 & Info\footnotemark[3] & be back & 0.17 & be & part & 0.17 \\
we & have & 2.24 & be & who & 0.88 & I & be & 0.17 & take & place & 0.10 \\
you & see & 2.16 & say & what & 0.88 & which & be & 0.14 & be & high & 0.14 \\
you & let & 1.71 & take & place & 0.77 & you & can follow & 0.12 & can follow & any response & 0.12 \\
I & know & 1.65 & tell & I & 0.75 & platinum\footnotemark[4] & be back & 0.12 & order today & today & 0.12 \\
I & have & 1.57 & be & in fact & 0.72 & you & order today & 0.12 & be & safe & 0.12 \\
you & know & 1.49 & be & right & 0.67 & this entry & file & 0.12 & order today & free shipping & 0.12 \\
we & know & 1.46 & tell & we & 0.66 & this entry & post & 0.12 & order today & vitamin\footnotemark[5] & 0.12 \\
you & have & 1.42 & know & what & 0.60 & you & check & 0.11 & say & in statement & 0.12 \\
what & be & 1.37 & call & what & 0.59 & you & have & 0.11 & order today & with\footnotemark[2] & 0.12 \\
what & happen & 1.33 & be &  & 0.58 & you & read & 0.11 & order today & at \% & 0.12 \\
I & believe & 1.23 & be & time & 0.57 & you & note & 0.11 & be & responsible & 0.12 \\
people & be & 1.21 & be & responsible & 0.55 & study & publish & 0.10 & be & important & 0.11 \\
I & say & 1.20 & be & possible & 0.51 & fluoride & be & 0.10 & can follow & through\footnotemark[6] & 0.09 \\ \bottomrule
\end{tabular}
\begin{tabular}{@{}l@{}}
 \footnotemark[1] circulate today \\
 \footnotemark[2] with double Patriot Points \\
 \footnotemark[3] Infowars Life Lung Cleanse Plus \\
 \footnotemark[4] platinum standard \\
 \footnotemark[5] Vitamin Mineral Fusion \\
 \footnotemark[6] through rss feed
\end{tabular}
\end{table}

Analysis of top ``Agent-Event" edge frequencies in high-conspiracy documents reveals a prominent role for first and second-person pronouns (``I", ``we", ``you"), often coupled with cognitive verbs (e.g., ``think" and ``believe"). 
To provide a more comprehensive picture, we extended the analysis to the top 100 edges; however, Although Table \ref{tab:edges_freq} reports only the top 20,  the extended analysis shows pattern where these pronouns most often pair with verbs such as ``think", ``do", ``believe", ``get", ``make", and ``find". These verb pairings suggest distinct narrative functions. Connections involving verbs like ``think" and ``believe" relate to cognitive processes and belief formation. Verbs such as ``do", ``get", ``make", and ``find" imply a focus on actions and agency within the narrative. Notably, the verbs ``be", ``have", ``know", and ``see" also show frequent connections with ``I", ``we", and ``you". However, we did not interpret their implications for conspiracy narratives because they appear with equally high edge frequencies in low-similarity conspiracy documents.

\end{document}